\pdfoutput=1

\documentclass[11pt]{article}

\usepackage{EMNLP2023}

\usepackage{times}
\usepackage{latexsym}
\usepackage{graphicx}

\usepackage[T1]{fontenc}

\usepackage[utf8]{inputenc}

\usepackage{microtype}

\usepackage{inconsolata}

\usepackage{algorithm}
\usepackage{algpseudocode}
\usepackage{amsmath}
\usepackage{tikz}
\usepackage{pgf-pie}
\usepackage{multirow}
\usepackage{tabularx}
\usepackage{booktabs}
\usepackage[frozencache=true,cachedir=./]{minted}
\usepackage{multicol}
\usepackage{listings}

\title{Prompting with Pseudo-Code Instructions}

\author{\bf{Mayank Mishra\thanks{\hspace{1ex}Equal contribution}, Prince Kumar\footnotemark[1], Riyaz Bhat,} \\
\bf{Rudra Murthy V, Danish Contractor, Srikanth Tamilselvam}\\
IBM Research AI \\
\{mayank.mishra1,prince.kumar12, riyaz.bhat,danish.contractor\}@ibm.com,
\\
\{rmurthyv,srikanth.tamilselvam\}@in.ibm.com
}

\begin{document}
\maketitle
\begin{abstract}
Prompting with natural language instructions has recently emerged as a popular method of harnessing the capabilities of large language models (LLM). Given the inherent ambiguity present in natural language, it is intuitive to consider the possible advantages of prompting with less ambiguous prompt styles, like pseudo-code. 

In this paper, we explore if prompting via pseudo-code instructions helps improve the performance of pre-trained language models.  We manually create a dataset\footnote{Code and dataset available at \url{https://github.com/mayank31398/pseudo-code-instructions}} of pseudo-code prompts for $132$ different tasks spanning classification, QA, and generative language tasks, sourced from the Super-NaturalInstructions dataset \cite{wang-etal-2022-super}. Using these prompts along with their counterparts in natural language, we study their performance on two LLM families - BLOOM \cite{BLOOM}, CodeGen \cite{codegen}. Our experiments show that using pseudo-code instructions leads to better results, with an average increase (absolute) of 7-16 points in F1 scores for classification tasks and an improvement (relative) of 12-38\% in aggregate ROUGE-L scores across all tasks. We include detailed ablation studies which indicate that code comments, docstrings, and the structural clues encoded in pseudo-code all contribute towards the improvement in performance. 
To the best of our knowledge, our work is the first to demonstrate how pseudo-code prompts can be helpful in improving the performance of pre-trained LMs.
\end{abstract}

\section{Introduction}

\begin{listing}[!htb]
\begin{minted}{python}
def generate_sentiment(sentence: str) -> str:
    """For the given sentence, the task is to
    predict the sentiment. For positive 
    sentiment return "positive" else return
    "negative".

    Parameters:
        sentence (str): input sentence
    Returns:
        str: sentiment of the input
    """

    # predict the sentiment
    if sentiment_is_positive(sentence):
        return "positive"
    else:
        return "negative"

>>> generate_sentiment(
    "that has a charmingly bourbon air."
)
\end{minted}
\caption{An example pseudo-code instruction for the task from \citet{wang-etal-2022-super}. A successful model is expected to use the provided pseudo-code instructions and output responses to a pool of evaluation instances.}
\label{fig:schema}
\end{listing}

Prompting with natural language instructions has recently emerged as a popular method of harnessing the capabilities of large language models. In addition to fine-tuning, models are often fine-tuned using instructions on a large collection of datasets to help improve the ability of LMs to follow instructions and performance on unseen tasks \cite{wei2022finetuned,wang-etal-2022-super}. 

However, natural language instructions can be ambiguous and under-specified, and therefore have multiple interpretations -- including detailed instructions may not always be beneficial, as it can add to the complexity of reasoning for models. This has led to the growing body of work around `prompt-engineering'  where specialized prompting strategies are developed for different domains and task types \cite{pmlr-v139-zhao21c,10.1145/3411763.3451760, arora2023ask, 10.1145/3560815, 10.1145/3544548.3581388}. In addition, inference-time prompting strategies that specifically aid multi-step reasoning have also been found to be helpful -- e.g: the inclusion of chain-of-thought reasoning in few-shot settings results in improved performance over standard prompts \cite{cot-few-shot},  the infamous ``{\em Let's think step-by-step''}-prompt for boosting 0-shot performance \cite{stepbystep}. 
\begin{algorithm}[]
\scriptsize
\caption{Attention Block}\label{alg:transformersattention}
\begin{algorithmic}[1]
\Function{transformers\_attention\_block}{$Q$, $K$, $V$}
    \State \textbf{Input:} $Q$, $K$, and $V$: input matrices.
    \State \textbf{Output:} The output of the attention block.
    \State $scores \gets Q \cdot K^T$
    \State $attention\_weights \gets softmax(scores)$
    \State $weighted\_values \gets attention\_weights \cdot V$
    \State $output \gets \sum_{i=1}^{n} weighted\_values_i$
    \State \textbf{return} $output$
\EndFunction
\end{algorithmic}
\end{algorithm}

Given the inherent ambiguity present in natural language, it is intuitive to consider the advantages of prompting with less ambiguous prompt styles, such as the use of pseudo-code.
Pseudo-code is an informal set of code-like constructs, which tend to be easy to interpret for humans but are not necessarily compilable/executable. They are often used to express complex ideas, processes, and flows -- for example, Algorithm \ref{alg:transformersattention} expresses a summarized version of what happens within a Multi-Head Attention block \cite{vaswani2017attention} in pseudo-code. Arguably, expressing the same ideas in natural language could result in ambiguity and would perhaps require detailed text for clarity, which adds to the complexity. 

In light of recent successes in NLP tasks achieved by code models \cite{madaan2022language, zhang2023exploring, zhang2023causal}, this study aims to examine the efficacy of using pseudo-code instructions for prompting as a means of enhancing model performance. This study is driven by the hypothesis that using pseudo-code as prompts could offer a natural advantage to models in NLP tasks, owing to the concise and clearer expression of ideas in pseudo-code. To test the hypothesis that prompting large language models with pseudo-code instead of natural language data could be helpful, we created pseudo-code prompts\footnote{The pseudo-code instructions for each of these tasks were created by the authors of this paper.} for $132$ different tasks spanning 28 distinct task types, sourced from the Super-NaturalInstructions dataset \cite{wang-etal-2022-super} (see Listing \ref{fig:schema} for an example).
Using these prompts along with their counterparts from natural language, we study their performance on two LLM families: BLOOM \cite{BLOOM} and CodeGen \cite{codegen}. Both LLM families have been trained on natural language as well as code data.

We compare the performance of both styles of prompts on classification tasks, QA tasks, as well as a mix of other language generation tasks. Our experiments indicate that prompting with pseudo-code instructions indeed helps, and they result in an absolute gain of 7-16 points in F1 scores on classification tasks, and 12-38\% relative improvement in aggregate ROUGE-L scores across all tasks.

\noindent{\paragraph{Contributions:}
In summary, our paper makes the following contributions: 
(i) We release a dataset of $132$ pseudo-code prompts spanning 28 different task types; (ii)
Through a series of detailed experiments on two publicly available open-access LLM families, we demonstrate how prompting with pseudo-code instructions results in a marked improvement in performance over prompting with natural language instructions;
(iii) We include detailed ablation studies indicating that code comments, docstrings, and the structural clues encoded in pseudo-code all contribute towards the improvement in performance.

To the best of our knowledge, our work is the first to demonstrate how pseudo-code prompts\footnote{In the rest of the paper, we use the words `pseudo-code' and `code' interchangeably when referring to prompts.} can be helpful in improving the performance of pre-trained LMs. Our findings not only emphasize the significance of leveraging pseudo-code for prompting but also shed light on the specific elements within pseudo-code that contribute to the observed improvements.

\section{Related Work}
Finetuning large language models on instruction datasets can enhance their performance and even their ability to generalize to unseen tasks \cite{wei2021finetuned,chung2022scaling}. Many aspects of instruction finetuning such as the number of tasks, model size, and finetuning on chain-of-thought data have been found to be useful \cite{chung2022scaling}. Consequently, significant efforts have been invested in manually creating instruction datasets, as well as using existing generative models to train and evaluate language models \cite{mishra2021cross, bach2022promptsource, wang-etal-2022-super, wang2022self}. The instructions available in instruction tuning datasets are mostly in natural language, but have been applied for both natural language tasks and programming tasks. But alternatives to natural language instructions such as programming language code, pseudo-code, symbols \cite{maccartney2007natural} etc. have not been thoroughly explored even for programming tasks. Compared to natural language, code or pseudo-code has less ambiguity due to its inherent nature of using functions or steps that contribute towards accomplishing a task. This makes them a natural choice for specifying instructions. 
Recently, few works \cite{askmarvin, madaan2022language, zhang2023exploring, zhang2023causal} have explored code and pseudo-code as inputs. Unlike contemporaneous work by \citet{zhang2023exploring} we find that pseudo-code instructions indeed provide better performance over NL instructions on a wide variety of tasks. 

\section{Dataset}
The Super-NaturalInstructions dataset \cite{wang-etal-2022-super} comprises $1,616$ diverse NLP tasks, and each task contains the task instruction, positive/negative examples, and instances. We sampled a mixture of $132$ tasks that did not require multilingual capabilities and re-wrote instructions for a subset of this dataset using Python constructs. Note that we borrow Python constructs only to express our prompts in pseudo-code and our prompts do not result in executable Python code. Further, we do not include any additional steps/instructions that were not present in the original natural language instructions.

All task instructions follow the schema as described in Listing \ref{fig:schema}. The schema consists of the following elements.

    \noindent \paragraph{Function Prototype:} This defines the prototype of the main pseudo-code function. The function names are descriptive and summarize the task to be performed. They also include all variables passed as input along with their data types and return type.  We follow the PEP 8\footnote{\url{https://peps.python.org/pep-0008/}} style guidelines for writing the pseudo-code and use strongly typed prototypes. We avoid declaring global variables whenever possible and pass them as arguments to a method. To the extent possible, we also avoid the use of classes and enumerations. Line number $1$ in Listing \ref{fig:schema} provides an example function prototype for a sentiment classification task. 
    
    \noindent \paragraph{DocString:} The docstring provides detailed instructions on the task to be performed in natural language. Often, this is a paraphrased version of the original natural language instruction. The docstring ends with a list of parameters (with their types) being passed and the return type from the function. An example docstring for the sentiment classification task is presented in line numbers 2 to 12 in Listing \ref{fig:schema}.

    \noindent \paragraph{Function Definition:} This includes the bulk of the pseudo-code instruction describing how to solve the particular task. To the extent possible, the function definitions do not leave out any information contained in the docstring. Pseudo-code in the function definition are written as sub-task functions. These sub-task functions are usually not defined and often use descriptive names, arguments and variables. We include in-line comments indicating what is accomplished by the sub-task function and the role of the arguments if required. We sometimes also define secondary sub-task functions if it requires additional details or if the descriptive function name may not be adequate to specify the goal of the sub-task function. We assume the availability of basic helper functions such as \texttt{concat\_str}, \texttt{search} etc., and do not include any import statements.
    
    Line numbers 13 to 16 present function definition for sentiment classification task. The function calls \texttt{sentiment\_is\_positive} sub-task function which checks if the sentiment of the given sentence is positive or not. This function is not explicitly defined in the instruction.
    
    \noindent{\paragraph{Pre-processor:} Since the pseudo-code instructions expect inputs as arguments, we need to parse the inputs provided in the Super-NaturalInstructions dataset \cite{wang-etal-2022-super} (which provides pre-formatted inputs). For each pseudo-code instruction, we also include an executable python pre-processor which is used for parsing the input.

\subsection{Dataset Statistics}
We created instructions for $132$ tasks that have instructions and input/output pairs in English language. We group the tasks into three classes: Classification Tasks (Table \ref{tab:classification_dataset_stats}), QA tasks (Table \ref{tab:qa_dataset_stats}) and other language generation tasks (Table \ref{tab:gen_dataset_stats}). These tasks cover a total of $28$ different categories and span 72 unique datasets. For each task we sample $1000$ instances for evaluation.

\section{Evaluation}
In order to study if instruction specification via pseudo-code results in improved performance over baseline NL English instructions, we choose to experiment with BLOOM \cite{BLOOM}, CodeGen \cite{codegen} models.
Our choice of models is motivated by the fact that these models have not been instruction-fine-tuned on the Natural Instructions dataset. In addition, they have both been trained on code and natural language data. 

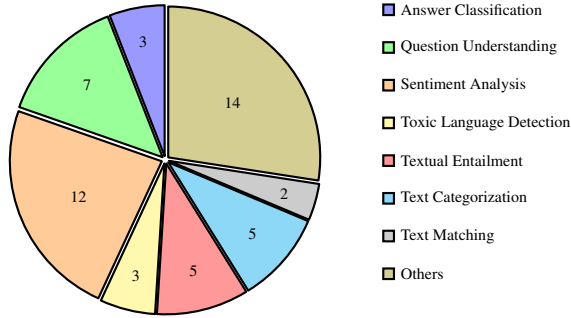
\begin{figure}[H]
\centering
{\tiny
\begin{tikzpicture}
\pie[radius=2, explode=.05, color={blue!40, green!40, orange!40, yellow!40, red!40, cyan!40, gray!40, olive!40}, sum=auto, text=legend, rotate=90]
  {
    3/Answer Classification,
    7/Question Understanding,
    12/Sentiment Analysis,
    3/Toxic Language Detection,
    5/Textual Entailment,
    5/Text Categorization,
    2/Text Matching,
    14/Others
  }
\end{tikzpicture}}

\bigskip

{\tiny
\begin{tabular}{p{2.5cm}|p{4.3cm}}
\textbf{Task Category} & \textbf{Datasets} \\
\hline
Answer Classification & MultiRC \cite{khashabi-etal-2018-looking}, McTaco \cite{ZKNR19}, TWEETQA \cite{xiong-etal-2019-tweetqa} \\
Answer Verification & MultiRC \cite{khashabi-etal-2018-looking} \\
Commonsense Classification & ATOMIC \cite{Sap2019ATOMICAA} \\
Coreference Selection & Numeric Fused-Head \cite{elazar-goldberg-2019-wheres} \\
Dialogue Selection & SPOLIN \cite{cho2020spolin}, DSTC3 \cite{henderson2014third} \\
Grammar Error Detection & CoLA \cite{warstadt-etal-2019-neural} \\
Intent Identification & DailyDialog \cite{li-etal-2017-dailydialog} \\
Irony Detection & SemEval2018-Task3 \cite{van-hee-etal-2018-semeval} \\
Linguistic Classification & SentEval \cite{conneau-kiela-2018-senteval} \\
Prime Number Classification & Synthetic \cite{wang-etal-2022-super} \\
Program Execution & Synthetic \cite{wang-etal-2022-super} \\
Question Understanding & McTaco \cite{ZKNR19}, DROP \cite{Dua2019DROPAR}, TREC \cite{li-roth-2002-learning}, DREAM \cite{sun-etal-2019-dream}, FreebaseQA \cite{jiang-etal-2019-freebaseqa} \\
Section Classification & CODA-19 \cite{huang-etal-2020-coda} \\
Sentiment Analysis & The Multilingual Amazon Reviews Corpus \cite{keung-etal-2020-multilingual}, Sentiment140 \cite{go2009twitter}, SST-2 \cite{socher-etal-2013-recursive}, PerSenT \cite{bastan2020authors}, Amazon Review Polarity \cite{amazon-polarity-dataset}, PEC \cite{zhong-etal-2020-towards}, Poem Sentiment \cite{sheng-uthus-2020-investigating} \\
Text Categorization & MultiNLI \cite{williams-etal-2018-broad}, DDO \cite{durmus-cardie-2019-corpus}, SemEval-2020 Task 7 \cite{hossain-etal-2020-semeval}, Scruples \cite{Lourie_Le_Bras_Choi_2021} \\
Text Matching & AFS \cite{misra-etal-2016-measuring}, PAWS \cite{paws2019naacl} \\
Text Quality Classification & McTaco \cite{ZKNR19} \\
Textual Entailment & MultiNLI \cite{williams-etal-2018-broad}, SNLI \cite{bowman-etal-2015-large}, e-SNLI \cite{NIPS2018_8163}, Defeasible-NLI \cite{rudinger-etal-2020-thinking}, ATOMIC \cite{Sap2019ATOMICAA} \\
Toxic Language Detection & CAD \cite{vidgen-etal-2021-introducing}, Jigsaw \cite{jigsaw-unintended-bias-in-toxicity-classification}, Hate Speech Offensive \cite{Davidson_Warmsley_Macy_Weber_2017} \\
Wrong Candidate Generation & McTaco \cite{ZKNR19} \\
\end{tabular}}
\captionof{table}{Collection of classification tasks used in our work}
\label{tab:classification_dataset_stats}
\end{figure}

\begin{figure}[!ht]
\bigskip
\centering
{\tiny
\begin{tikzpicture}
\pie[radius=2, explode=0.05, color={blue!40, green!40, orange!40}, sum=auto, text=legend, rotate=90]
{
    10/Extractive QA,
    16/Generative QA,
    8/MCQ
}
\end{tikzpicture}}

\bigskip
{\tiny
\begin{tabular}{p{1.5cm}|p{5.3cm}}
\textbf{Task Category} & \textbf{Datasets} \\
\hline
Extractive QA & ROPES \cite{Lin2019ReasoningOP}, Odd-Man-Out \cite{Stanovsky2016EMNLP}, SQuAD1.1 \cite{rajpurkar-etal-2016-squad}, Synthetic \cite{wang-etal-2022-super}, MCScript \cite{ostermann-etal-2018-mcscript}, PICO \cite{jin-szolovits-2018-pico}, MWSC \cite{mccann2019the}, OPUS \cite{TIEDEMANN12.463}, CoQA \cite{reddy-etal-2019-coqa} \\
Generative QA & Quoref \cite{dasigi-etal-2019-quoref}, McTaco \cite{ZKNR19}, DROP \cite{Dua2019DROPAR}, MultiRC \cite{khashabi-etal-2018-looking}, PIQA \cite{Bisk2020}, Synthetic \cite{wang-etal-2022-super}, BREAK \cite{Wolfson2020Break}, Natural Questions \cite{47761}, AmbigQA \cite{min-etal-2020-ambigqa}, CoQA \cite{reddy-etal-2019-coqa}, TriviaQA \cite{joshi-etal-2017-triviaqa} \\
MCQ & Essential, QuaRel \cite{Tafjord2018QuaRelAD}, WinoGrande \cite{sakaguchi2021winogrande}, MultiNLI \cite{williams-etal-2018-broad}, ReCoRD \cite{Zhang2018ReCoRDBT}, MMMLU \cite{hendryckstest2021} \\
\end{tabular}}
\captionof{table}{Collection of QA tasks used in our work}
\label{tab:qa_dataset_stats}
\end{figure}

\begin{figure}[!ht]
\centering
{\tiny
\begin{tikzpicture}
\pie[radius=2, explode=0.05, color={blue!40, green!40, orange!40, yellow!40, red!40, cyan!40}, sum=auto, text=legend, rotate=90]
{
    6/List Operation,
    3/Option Generation,
    2/Paraphrasing,
    11/Question Generation,
    10/Rewriting,
    15/Misc.
}
\end{tikzpicture}}

\bigskip
{\tiny
\begin{tabular}{p{1.8cm}|p{5cm}}
\textbf{Task Category} & \textbf{Datasets} \\
\hline
List Operation & CoNaLa \cite{yin2018mining}, Synthetic \cite{TIEDEMANN12.463}, Youtube Caption Corrections \cite{YTCC} \\
Option Generation & aNLI \cite{nie2019adversarial}, ASSET \cite{alva-manchego-etal-2020-asset}, ROCStories \cite{mostafazadeh2017lsdsem} \\
Paraphrasing & ZEST \cite{weller-etal-2020-learning}, PARANMT-50M \cite{wieting-gimpel-2018-paranmt} \\
Question Generation & CosmosQA \cite{huang-etal-2019-cosmos}, WinoGrande \cite{sakaguchi2021winogrande}, ROPES \cite{lin-etal-2019-reasoning}, SQuAD1.1 \cite{rajpurkar-etal-2016-squad}, StrategyQA \cite{geva-etal-2021-aristotle}, SQuAD2.0 \cite{rajpurkar-etal-2018-know}, BoolQ \cite{clark-etal-2019-boolq}, CoQA \cite{reddy-etal-2019-coqa}, QA-ZRE \cite{levy-etal-2017-zero} \\
Rewriting & WinoGrande \cite{sakaguchi2021winogrande}, aNLI \cite{nie2019adversarial}, ASSET \cite{alva-manchego-etal-2020-asset}, ZEST \cite{weller-etal-2020-learning}, SNLI \cite{bowman-etal-2015-large} \\
Misc. & DROP \cite{Dua2019DROPAR}, WinoGrande \cite{sakaguchi2021winogrande}, QASC \cite{Khot_Clark_Guerquin_Jansen_Sabharwal_2020}, Essential \cite{khashabi-etal-2017-learning}, ROPES \cite{Lin2019ReasoningOP}, StoryCloze \cite{mostafazadeh-etal-2016-corpus}, Country Barcode Prefix dataset, Country Region in World dataset, Gigaword \cite{graff2003english}, GAP \cite{webster-etal-2018-mind}, SPOLIN \cite{cho2020spolin}, XL-WiC \cite{raganato-etal-2020-xl} \\
\end{tabular}}
\captionof{table}{Collection of language generation tasks used in our work}
\label{tab:gen_dataset_stats}
\end{figure}

The BLOOM models are trained on the ROOTS corpus \cite{ROOTS} consisting of 46 natural and 13 programming languages. On the other hand, the CodeGen models are trained on the Pile corpus \cite{gao2020pile}, Google's publicly available BigQuery and BigPython datasets~\cite{codegen}. The BLOOM models have been trained on a mixture of natural language and code simultaneously. As for the CodeGen models we utilize, they were initially trained on natural language and subsequently received additional training focused specifically on Python code.

Our choice of models allows us to setup a controlled environment where we can study the impact of prompting in natural language and pseudo-code. Most recent instruction-tuned models have either seen the Super-NaturalInstructions dataset \cite{wang-etal-2022-super} in some form \cite{longpre2023flan} or they do not have tokenizers that will meaningfully process code syntax \cite{raffel2020exploring}, and therefore can not be used in our study. By empirically studying the performance of models on these prompts, we hope to inform future work on training an instruction-tuned model using pseudo-code instructions. 

\begin{table*}[ht]
\centering
\fontsize{6.8}{7}\selectfont
\begin{tabular}{cc|ccc|c|c|ccc}
\toprule
\textbf{Model} & \textbf{\begin{tabular}[c]{@{}c@{}}Instruction\\  Format\end{tabular}} & \multicolumn{3}{c|}{\textbf{Classification Tasks}}      & \textbf{QA Tasks} & \textbf{\begin{tabular}[c]{@{}c@{}}Generation tasks\end{tabular}} & \multicolumn{3}{c}{\textbf{All Tasks}} \\ \midrule
      & & \multicolumn{1}{c}{Macro F1}       & \multicolumn{1}{c}{Micro F1}       & Weighted F1    & ROUGE-L    & ROUGE-L   & \multicolumn{1}{c}{ROUGE-L} & \multicolumn{1}{c}{ANLS}  & EM    \\ \midrule
\multicolumn{1}{c}{Majority Class} &  & \textbf{0.296} & \textbf{0.509} & \textbf{0.362} & - & - & - & - & - \\ \midrule
\multirow{2}{*}{CodeGen 2B} & \multicolumn{1}{c|}{Code Instructions} &    \textbf{0.272}  &    \textbf{0.417}   &    \textbf{0.354}   &    \textbf{0.175}  &    \textbf{0.317}   &    \textbf{0.330}  &  \textbf{0.261}   &    \textbf{0.202}  \\ 
\cmidrule(lr){2-10}
& \multicolumn{1}{c|}{NL Instructions} & \multicolumn{1}{c}{0.068}           &    \multicolumn{1}{c}{0.306}   &    {0.239}   &    {0.154}   &    0.254     &    \multicolumn{1}{c}{0.265}   &    \multicolumn{1}{c}{0.195}   &    {0.147} \\ \midrule

\multirow{2}{*}{CodeGen 6B} & \multicolumn{1}{c|}{Code Instructions} & \textbf{0.311} & \textbf{0.443} & \textbf{0.375} & \textbf{0.201} & \textbf{0.327} & \textbf{0.354} & \textbf{0.283} & \textbf{0.218} \\  \cmidrule(lr){2-10}
& \multicolumn{1}{c|}{NL Instructions} & \multicolumn{1}{c}{0.052}  & \multicolumn{1}{c}{0.278}  & 0.215  & 0.132   & 0.271              & \multicolumn{1}{c}{0.257}  & \multicolumn{1}{c}{0.187}  & 0.134  \\ \midrule

\multirow{2}{*}{BLOOM 3B} & \multicolumn{1}{c|}{Code Instructions} & \textbf{0.116} & \textbf{0.351} & \textbf{0.288} & 0.147 & \textbf{0.271} & \textbf{0.279} & \textbf{0.215} & \textbf{0.165} \\ \cmidrule(lr){2-10}
& \multicolumn{1}{c|}{NL Instructions} & \multicolumn{1}{c}{0.082}  & \multicolumn{1}{c}{0.275}  & 0.214  & \textbf{0.159} & 0.234       & \multicolumn{1}{c}{0.250}  & \multicolumn{1}{c}{0.180}  & 0.132  \\ \midrule

\multirow{2}{*}{BLOOM 7B} & \multicolumn{1}{c|}{Code Instructions} & \textbf{0.174} & \textbf{0.369} & \textbf{0.285} & 0.150 & \textbf{0.298} & \textbf{0.297} & \textbf{0.232} & \textbf{0.176} \\ \cmidrule(lr){2-10}
& \multicolumn{1}{c|}{NL Instructions} & \multicolumn{1}{c}{0.046}  & \multicolumn{1}{c}{0.247}  & 0.203  & \textbf{0.156}   & 0.276       & \multicolumn{1}{c}{0.247}  & \multicolumn{1}{c}{0.172}  & 0.122  \\ 


\bottomrule
\end{tabular}
\caption{Performance of models when prompted using pseudo-code instructions and natural language instructions in 0-shot settings. (i) In each model, prompting with pseudo-code instructions results in much higher performance in almost all the tasks (ii) For each model family, increasing scale helps improve performance (iii) Prompting CodeGen (a model designed for code) results in better performance than BLOOM. (iv) Prompting BLOOM models with Natural Language instructions instead of code-instructions results in higher performance on QA tasks.} \label{tab:zero-shot-main}
\end{table*}

\subsection{Model Configurations}
For all of the experiments conducted in this paper, we use BLOOM-3B, BLOOM 7B \cite{BLOOM}, CodeGen-mono 2B, and CodeGen-mono 6B \cite{codegen}  models. The inference was performed using A100 80 GB GPUs. To accelerate the inference of all models, we utilized DeepSpeed-Inference \cite{aminabadi2022deepspeed} in fp16, which resulted in an average inference throughput improvement of around 1.7x, compared to the standard HuggingFace \cite{wolf-etal-2020-transformers} inference. We used greedy decoding for all our experiments for reproducibility and restricted generated outputs to $100$ tokens. Even for classification tasks, we generate the class labels using auto-regressive decoding instead of picking the class label with lowest perplexity. This is done because not all class labels can be mapped to a single token for all tasks. This technique of evaluating performance of classification tasks is often employed when  using closed LLMs, such as those behind APIs (eg: OpenAI's GPT4 \cite{openai2023gpt4}, Google's PaLM \cite{chowdhery2022palm} etc).

\subsection{Metrics}
We adopt different metrics for each task-category: we measure the performance of classification tasks using micro, macro and weighted F1 scores, and for QA and language generation tasks we use the ROUGE-L metric. We report the ROUGE-L, Exact Match (EM), and ANLS - Average Normalized Levenshtein Similarity \cite{biten2019scene} for all tasks. 

\newcolumntype{L}{>{\raggedright\arraybackslash}X}

\subsection{Output post-processing} Since the models we experiment with have not been fine-tuned for instruction following, they tend to generate excess text after the output for the given task. We therefore post-process the outputs to ensure models are not penalized in our evaluation due to excess generations. We post-process all outputs by truncating by the newline character \verb|'\n'|. Furthermore, the output is subjected to additional post-processing, including punctuation removal and lower casing. 

\subsection{Results}
Through our experiments we aim to answer the following questions: (i) What is the difference in performance between prompting pre-trained language and code models with pseudo-code prompts versus natural language prompts? (ii) How does increasing model size affect the efficacy of pseudo-code prompts? (iii) To what extent does structured prompting, such as the use of function names, docstrings, inline comments, and arguments, impact performance on tasks? 

\subsubsection{Prompting with Pseudo-code}
Table \ref{tab:zero-shot-main} compares the performance of prompting with pseudo-code (referred to as code instructions) and natural language instructions in 0-shot settings. Results have been grouped by model family and size. 

As can be seen, for all model families and sizes, prompting with pseudo-code results in a significant improvement in performance. The performance on classification tasks is especially notable, for example, the gains on weighted F1 vary between 7-16 F1 points (absolute). Furthermore, the relative performance improvement on all other tasks, as measured by ROUGE-L, varies between 12-38\%. The overall performance as measured by ROUGE-L, ANLS and Exact Match also report similar trends. 

\noindent{\paragraph{Comparison of CodeGen vs BLOOM} Despite most tasks being non-code tasks, CodeGen, a model designed for code applications, outperforms BLOOM models, even when using natural language instructions (see metrics for ‘All Tasks’). Similar behavior has been anecdotally reported \cite{fu2022gptroadmap, madaan2022language}, but has possibly not been investigated using as many tasks as presented in this paper. Note, however, that using pseudo-code prompts in the code models results in better performance than any other prompt-model configuration.

\begin{table*}[!htb]
\scriptsize
\begin{tabular}{cccc|ccc|ccc|ccc}
\toprule
 & \multicolumn{6}{c|}{\textbf{CodeGen 6B}}       & \multicolumn{6}{c}{\textbf{BLOOM 7B}} \\ 
 \cmidrule(lr){2-13} 
  & \multicolumn{3}{c|}{\textbf{Code Instructions}} & \multicolumn{3}{c|}{\textbf{NL Instructions}}   & \multicolumn{3}{c|}{\textbf{Code Instructions}} & \multicolumn{3}{c}{\textbf{NL Instructions}} \\ 
  \midrule
\multicolumn{1}{c}{QA Task} & \multicolumn{1}{c}{EM} & \multicolumn{1}{c}{ROUGE-L} & \multicolumn{1}{c|}{ANLS} & \multicolumn{1}{c}{EM} & \multicolumn{1}{c}{ROUGE-L} & \multicolumn{1}{c|}{ANLS} & \multicolumn{1}{c}{EM} & \multicolumn{1}{c}{ROUGE-L} & \multicolumn{1}{c|}{ANLS} & \multicolumn{1}{c}{EM} & \multicolumn{1}{c}{ROUGE-L} & \multicolumn{1}{c}{ANLS} \\ \midrule
\multicolumn{1}{c}{\textbf{Extractive QA}} & 0.140  & 0.303 & 0.189 & 0.045 & 0.188 & 0.077 & 0.047 & 0.184 & 0.077 & 0.047 & 0.227 & 0.086 \\ 
\midrule
\multicolumn{1}{c}{\textbf{Generative QA}} & 0.045 & 0.129 & 0.068 & 0.029 & 0.095 & 0.045 & 0.028    & 0.101 & 0.042 & 0.032 & 0.115 & 0.047         \\ \midrule
\multicolumn{1}{c}{\textbf{MCQ}}  & 0.196 & 0.213 & 0.210 & 0.082 & 0.106 & 0.083 & 0.184 & 0.201 & 0.197 & 0.107 & \multicolumn{1}{c}{0.143} & 0.108         \\ 
\bottomrule
\end{tabular}
\caption{0-shot performance of CodeGen 6B and BLOOM 7B models on QA tasks from our dataset. As can be seen, pseudo-code instructions applied on the CodeGen model results in the best overall performance on all categories of QA tasks. However, comparing the performance of Natural Language Instructions, we find that it performs marginally  better than pseudo-code instructions on non-MCQ QA tasks when using the BLOOM 7B model.}
\label{tab:qa_details}
\end{table*}

\noindent{\paragraph{Performance on QA tasks} {\color{black}{Interestingly, we find that on QA tasks, the performance of pseudo-code instructions is better than natural-language instructions, when using the CodeGen model. However, this is not the case when using BLOOM.

We investigated this further and observed that for most QA tasks, the instructions in pseudo-code are not significantly more detailed or easier to understand than natural-language instructions. As an example, the pseudo-code instruction for answer generation from the SQuAD dataset merely contains the following statement in its function definition: \texttt{return get\_answer\_from\_passage(passage, question)} and reflects the details included in the natural instructions. 

We further analysed the results across QA task categories and found that pseudo-code instructions always help with multiple-choice questions (MCQ) tasks (see Table \ref{tab:qa_details} for a comparison between CodeGen 6B and BLOOM 7B).  We believe that this is because, understanding the instructions in such tasks may be more involved. For illustration, instructions in MCQ tasks often include details about {\em how} answers are expected -- eg: ``{\em choose the correct option A, B, C} '', ``{\em Select Option 1 - Value 1, Option 2 - Value 2} ''. Depending on the instructions, the models may be required to return options, values, or both which adds a degree of complexity to the instructions as compared to other types of QA.  

The discrepancy in performance between CodeGen and BLOOM on QA tasks (see Table \ref{tab:qa_details}), could be attributed to the fact that the structure from code prompts could be better leveraged by code models as programming languages and aspects of code syntax (structure) are likely to be better represented in a code model such as CodeGen. 
This brings us to our next question -- What is the contribution of structure that may be present in prompts?}}

\subsubsection{Contribution of Structure in prompts}

\begin{table*}[ht]
\fontsize{6.25}{7}\selectfont
\begin{tabular}{cccccccccc}
\toprule
\multicolumn{1}{c|}{\textbf{Model}} & \multicolumn{1}{c|}{\textbf{Instruction Format}}       & \multicolumn{3}{c|}{\textbf{Classification Tasks}} & \multicolumn{1}{c|}{\textbf{QA Tasks}} & \multicolumn{1}{c|}{\textbf{Generation Tasks}} & \multicolumn{3}{c}{\textbf{All Tasks}} \\ \midrule
\multicolumn{1}{c|}{}      & \multicolumn{1}{c|}{} & \multicolumn{1}{c|}{Macro F1}       & \multicolumn{1}{c|}{Micro F1}       & \multicolumn{1}{c|}{Weighted F1}    & \multicolumn{1}{c|}{ROUGE-L}    & \multicolumn{1}{c|}{ROUGE-L}  & \multicolumn{1}{c|}{ROUGE-L} & \multicolumn{1}{c|}{ANLS}  & EM    \\ \midrule
\multicolumn{1}{c|}{}      & \multicolumn{1}{c|}{Code Instructions (0)} & \multicolumn{1}{c|}{\textbf{0.272}} & \multicolumn{1}{c|}{\textbf{0.417}} & \multicolumn{1}{c|}{\textbf{0.354}} & \multicolumn{1}{c|}{0.175}    & \multicolumn{1}{c|}{\textbf{0.317}}   & \multicolumn{1}{c|}{\textbf{0.330}} & \multicolumn{1}{c|}{\textbf{0.262}} & \textbf{0.202} \\
\multicolumn{1}{c|}{}      & \multicolumn{1}{c|}{\begin{tabular}[c]{@{}c@{}}Function Declaration (0)\end{tabular}}       & \multicolumn{1}{c|}{0.159} & \multicolumn{1}{c|}{0.079} & \multicolumn{1}{c|}{0.085} & \multicolumn{1}{c|}{0.124}    & \multicolumn{1}{c|}{0.252}  & \multicolumn{1}{c|}{0.153} & \multicolumn{1}{c|}{0.083} & 0.043 \\
\multicolumn{1}{c|}{CodeGen 2B}     & \multicolumn{1}{c|}{\begin{tabular}[c]{@{}c@{}}Function Declaration (2)\end{tabular}}        & \multicolumn{1}{c|}{0.105} & \multicolumn{1}{c|}{0.267} & \multicolumn{1}{c|}{0.257} & \multicolumn{1}{c|}{\textbf{0.185}}    & \multicolumn{1}{c|}{0.294}  & \multicolumn{1}{c|}{0.256} & \multicolumn{1}{c|}{0.188} & 0.137 \\
\multicolumn{1}{c|}{}      & \multicolumn{1}{c|}{\begin{tabular}[c]{@{}c@{}}Function Invocation (2)\end{tabular}}         & \multicolumn{1}{c|}{0.097} & \multicolumn{1}{c|}{0.253} & \multicolumn{1}{c|}{0.238} & \multicolumn{1}{c|}{0.183}    & \multicolumn{1}{c|}{0.296}  & \multicolumn{1}{c|}{0.251} & \multicolumn{1}{c|}{0.183} & 0.131 \\
\multicolumn{1}{c|}{}      & \multicolumn{1}{c|}{\begin{tabular}[c]{@{}c@{}}Generic Function Invocation (2)\end{tabular}} & \multicolumn{1}{c|}{0.064} & \multicolumn{1}{c|}{0.282} & \multicolumn{1}{c|}{0.244} & \multicolumn{1}{c|}{0.167}    & \multicolumn{1}{c|}{0.257}  & \multicolumn{1}{c|}{0.245} & \multicolumn{1}{c|}{0.185} & 0.131 \\
\multicolumn{1}{l|}{}      & \multicolumn{1}{c|}{\begin{tabular}[c]{@{}c@{}}NL Examples (2)\end{tabular}}        & \multicolumn{1}{c|}{0.003}     & \multicolumn{1}{c|}{0.005}     & \multicolumn{1}{c|}{0.007}     & \multicolumn{1}{c|}{0.081}        & \multicolumn{1}{c|}{0.126}       & \multicolumn{1}{c|}{0.069}     & \multicolumn{1}{c|}{0.017}     & \multicolumn{1}{c}{0.006} \\ \midrule
\multicolumn{1}{c|}{}      & \multicolumn{1}{c|}{Code Instructions (0)} & \multicolumn{1}{c|}{\textbf{0.311}} & \multicolumn{1}{c|}{\textbf{0.444}} & \multicolumn{1}{c|}{\textbf{0.375}} & \multicolumn{1}{c|}{\textbf{0.201}}    & \multicolumn{1}{c|}{\textbf{0.327}}   & \multicolumn{1}{c|}{\textbf{0.354}} & \multicolumn{1}{c|}{\textbf{0.283}} & \textbf{0.218} \\
\multicolumn{1}{c|}{}      & \multicolumn{1}{c|}{\begin{tabular}[c]{@{}c@{}}Function Declaration (0)\end{tabular}}       & \multicolumn{1}{c|}{0.019} & \multicolumn{1}{c|}{0.101} & \multicolumn{1}{c|}{0.109} & \multicolumn{1}{c|}{0.162}    & \multicolumn{1}{c|}{0.273}  & \multicolumn{1}{c|}{0.179} & \multicolumn{1}{c|}{0.111} & 0.063 \\
\multicolumn{1}{c|}{CodeGen 6B}     & \multicolumn{1}{c|}{\begin{tabular}[c]{@{}c@{}}Function Declaration (2)\end{tabular}}        & \multicolumn{1}{c|}{0.134} & \multicolumn{1}{c|}{0.309} & \multicolumn{1}{c|}{0.281} & \multicolumn{1}{c|}{0.196}    & \multicolumn{1}{c|}{0.299}  & \multicolumn{1}{c|}{0.281} & \multicolumn{1}{c|}{0.212} & 0.154 \\
\multicolumn{1}{c|}{}      & \multicolumn{1}{c|}{\begin{tabular}[c]{@{}c@{}}Function Invocation (2)\end{tabular}}         & \multicolumn{1}{c|}{0.133} & \multicolumn{1}{c|}{0.296} & \multicolumn{1}{c|}{0.269} & \multicolumn{1}{c|}{0.192}    & \multicolumn{1}{c|}{0.302}  & \multicolumn{1}{c|}{0.275} & \multicolumn{1}{c|}{0.208} & 0.149 \\
\multicolumn{1}{c|}{}      & \multicolumn{1}{c|}{\begin{tabular}[c]{@{}c@{}}Generic Function Invocation (2)\end{tabular}} & \multicolumn{1}{c|}{0.062} & \multicolumn{1}{c|}{0.244} & \multicolumn{1}{c|}{0.215} & \multicolumn{1}{c|}{0.167}    & \multicolumn{1}{c|}{0.262}  & \multicolumn{1}{c|}{0.239} & \multicolumn{1}{c|}{0.175} & 0.121 \\
\multicolumn{1}{l|}{}      & \multicolumn{1}{c|}{\begin{tabular}[c]{@{}c@{}}NL Examples (2)\end{tabular}}        & \multicolumn{1}{c|}{0.000}     & \multicolumn{1}{c|}{0.000}     & \multicolumn{1}{c|}{0.001}     & \multicolumn{1}{c|}{0.102}        & \multicolumn{1}{c|}{0.168}       & \multicolumn{1}{c|}{0.088}     & \multicolumn{1}{c|}{0.023}    & \multicolumn{1}{c}{0.006} \\ \midrule
\multicolumn{1}{c|}{}      & \multicolumn{1}{c|}{Code Instructions (0)} & \multicolumn{1}{c|}{\textbf{0.116}} & \multicolumn{1}{c|}{\textbf{0.351}} & \multicolumn{1}{c|}{\textbf{0.288}} & \multicolumn{1}{c|}{0.147}    & \multicolumn{1}{c|}{\textbf{0.271}}   & \multicolumn{1}{c|}{\textbf{0.279}} & \multicolumn{1}{c|}{\textbf{0.214}} & \textbf{0.165} \\
\multicolumn{1}{c|}{}      & \multicolumn{1}{c|}{\begin{tabular}[c]{@{}c@{}}Function Declaration (0)\end{tabular}}       & \multicolumn{1}{c|}{0.000} & \multicolumn{1}{c|}{0.014} & \multicolumn{1}{c|}{0.016} & \multicolumn{1}{c|}{0.108}    & \multicolumn{1}{c|}{0.229}  & \multicolumn{1}{c|}{0.116} & \multicolumn{1}{c|}{0.054} & 0.015 \\
\multicolumn{1}{c|}{BLOOM 3B}       & \multicolumn{1}{c|}{\begin{tabular}[c]{@{}c@{}}Function Declaration (2)\end{tabular}}        & \multicolumn{1}{c|}{0.080} & \multicolumn{1}{c|}{0.237} & \multicolumn{1}{c|}{0.217} & \multicolumn{1}{c|}{\textbf{0.164}}    & \multicolumn{1}{c|}{0.249}  & \multicolumn{1}{c|}{0.225} & \multicolumn{1}{c|}{0.159} & 0.115 \\
\multicolumn{1}{c|}{}      & \multicolumn{1}{c|}{\begin{tabular}[c]{@{}c@{}}Function Invocation (2)\end{tabular}}         & \multicolumn{1}{c|}{0.073} & \multicolumn{1}{c|}{0.227} & \multicolumn{1}{c|}{0.211} & \multicolumn{1}{c|}{\textbf{0.164}}    & \multicolumn{1}{c|}{0.234}  & \multicolumn{1}{c|}{0.215} & \multicolumn{1}{c|}{0.149} & 0.107 \\
\multicolumn{1}{c|}{}      & \multicolumn{1}{c|}{\begin{tabular}[c]{@{}c@{}}Generic Function Invocation (2)\end{tabular}} & \multicolumn{1}{c|}{0.032} & \multicolumn{1}{c|}{0.173} & \multicolumn{1}{c|}{0.168} & \multicolumn{1}{c|}{0.161}    & \multicolumn{1}{c|}{0.246}  & \multicolumn{1}{c|}{0.203} & \multicolumn{1}{c|}{0.137} & 0.086 \\
\multicolumn{1}{c|}{}      & \multicolumn{1}{c|}{\begin{tabular}[c]{@{}c@{}}NL Examples (2)\end{tabular}}  & \multicolumn{1}{c|}{0.000}     & \multicolumn{1}{c|}{0.025}     & \multicolumn{1}{c|}{0.031}     & \multicolumn{1}{c|}{0.150}        & \multicolumn{1}{c|}{0.208}       & \multicolumn{1}{c|}{0.122}     & \multicolumn{1}{c|}{0.056}     & \multicolumn{1}{c}{0.024} \\ \midrule
\multicolumn{1}{c|}{}      & \multicolumn{1}{c|}{Code Instructions (0)} & \multicolumn{1}{c|}{\textbf{0.174}} & \multicolumn{1}{c|}{\textbf{0.369}} & \multicolumn{1}{c|}{\textbf{0.285}} & \multicolumn{1}{c|}{{0.150}}    & \multicolumn{1}{c|}{\textbf{0.298}}   & \multicolumn{1}{c|}{\textbf{0.297}} & \multicolumn{1}{c|}{\textbf{0.232}} & \textbf{0.176} \\
\multicolumn{1}{c|}{}      & \multicolumn{1}{c|}{\begin{tabular}[c]{@{}c@{}}Function Declaration (0)\end{tabular}}       & \multicolumn{1}{c|}{0.004} & \multicolumn{1}{c|}{0.021} & \multicolumn{1}{c|}{0.027} & \multicolumn{1}{c|}{0.111}    & \multicolumn{1}{c|}{0.242}  & \multicolumn{1}{c|}{0.124} & \multicolumn{1}{c|}{0.058} & 0.017 \\
\multicolumn{1}{c|}{BLOOM 7B}       & \multicolumn{1}{c|}{\begin{tabular}[c]{@{}c@{}}Function Declaration (2)\end{tabular}}        & \multicolumn{1}{c|}{0.072} & \multicolumn{1}{c|}{0.256} & \multicolumn{1}{c|}{0.227} & \multicolumn{1}{c|}{\textbf{0.191}}    & \multicolumn{1}{c|}{0.289}  & \multicolumn{1}{c|}{0.257} & \multicolumn{1}{c|}{0.182} & 0.128 \\
\multicolumn{1}{c|}{}      & \multicolumn{1}{c|}{\begin{tabular}[c]{@{}c@{}}Function Invocation (2)\end{tabular}}         & \multicolumn{1}{c|}{0.086} & \multicolumn{1}{c|}{0.248} & \multicolumn{1}{c|}{0.221} & \multicolumn{1}{c|}{0.189}    & \multicolumn{1}{c|}{0.286}  & \multicolumn{1}{c|}{0.250} & \multicolumn{1}{c|}{0.176} & 0.123 \\
\multicolumn{1}{c|}{}      & \multicolumn{1}{c|}{\begin{tabular}[c]{@{}c@{}}Generic Function Invocation (2)\end{tabular}} & \multicolumn{1}{c|}{0.039} & \multicolumn{1}{c|}{0.199} & \multicolumn{1}{c|}{0.178} & \multicolumn{1}{c|}{0.187}    & \multicolumn{1}{c|}{0.276}  & \multicolumn{1}{c|}{0.232} & \multicolumn{1}{c|}{0.155} & 0.097 \\
\multicolumn{1}{l|}{}      & \multicolumn{1}{c|}{\begin{tabular}[c]{@{}c@{}}NL Examples (2)\end{tabular}}        & \multicolumn{1}{c|}{0.000}     & \multicolumn{1}{c|}{0.009}     & \multicolumn{1}{c|}{0.009}     & \multicolumn{1}{c|}{0.132}        & \multicolumn{1}{c|}{0.182}       & \multicolumn{1}{c|}{0.106}     & \multicolumn{1}{c|}{0.038}     & \multicolumn{1}{c}{0.016} \\  \bottomrule
\end{tabular}
\caption{Study of structured prompts: Performance of models when prompted using 0-shot pseudo-code instructions, function declaration in 0-shot and 2-shot settings as well as 2-shot prompting with a `generic' function name and the use of only examples. The number N in the brackets indicates N-shot prompt. (i) Except for the performance on QA tasks, in each model, prompting with pseudo-code instructions results in much higher performance which indicates that detailed instructions are helpful (ii) For each model family, and prompting style, increasing model scale improves performance (iii) As before, prompting a model designed for code, CodeGen, results in better performance than BLOOM.} \label{tab:struct-prompt-main}
\end{table*}

The reasons behind the performance improvement when using pseudo-code prompts are likely to be a combination of factors, including the use of descriptive function names that convey the function's purpose (such as \texttt{get\_answer(question)}), a model that can effectively utilize structured information, and a structured prompt for a task that could further benefit from few-shot examples.

We therefore experiment with different structured prompting styles and report their results in Table \ref{tab:struct-prompt-main}. We study the performance of CodeGen and BLOOM with five types of prompts: (i) Pseudo-code instructions, (ii) Prompts that make use of function declaration (declare function name only), (iii) a structured prompt consisting only of task examples in 2-shot settings using the task-descriptive function name (iv) a structured prompt consisting only of task examples in 2-shot settings using a generic function name -- `func' (v) using the Natural Language examples (without instructions) in 2-shot settings. Details about each prompt have been included in the Appendix. 

We make three important observations from Table \ref{tab:struct-prompt-main}. First, code-instructions in 0-shot settings consistently yield the best overall performance compared to other structured prompts. Second, on average, the CodeGen model consistently outperforms BLOOM on all tasks. Lastly, the QA tasks in our dataset, which are relatively easy to express in natural language instructions, also benefit from structured prompts, particularly when prompted with examples.

It can be inferred from these observations that the performance gains resulting from the use of pseudo-code prompts are likely due to clearer task instructions, and not just the exploitation of superfluous patterns from in-context learning. These findings reinforce the results from the previous experiment, which showed that code models are more capable of exploiting structured prompts. In the case of QA tasks in our dataset, it is worth noting that since the pseudo-code instructions are not as detailed, even utilizing a simpler structured prompt with examples can significantly enhance performance as compared to natural language prompts.

\begin{table*}[ht]
\fontsize{6.5}{7}\selectfont
\centering
\begin{tabular}{cc|ccc|c|c|ccc}
\toprule
\multicolumn{1}{c}{\textbf{Model}} & \multicolumn{1}{c}{\textbf{Instruction Format}} & \multicolumn{3}{|c|}{\textbf{Classification Tasks}} & \multicolumn{1}{c|}{\textbf{QA Tasks}} & \multicolumn{1}{c|}{\textbf{Generation Tasks}} & \multicolumn{3}{c}{\textbf{All Tasks}} \\
\midrule
 &  & Macro F1 & Micro F1 & Weighted F1 & ROUGE-L & ROUGE-L & ROUGE-L & ANLS & EM\\
 \midrule
\multirow{2}{*}{CodeGen 6B} & \multicolumn{1}{c|}{Code Instructions} & \textbf{0.311} & \textbf{0.444} & \textbf{0.375} & \textbf{0.201} & \textbf{0.327} & \textbf{0.354} & \textbf{0.283} & \textbf{0.218}\\
\cmidrule(lr){2-10}
 & \begin{tabular}[c]{@{}c@{}}Code Instructions without\\ docstrings and comments\end{tabular} & 0.263 & 0.409 & 0.348 & 0.195 & 0.327 & 0.335 & 0.266 & 0.201\\
 \midrule
\multirow{2}{*}{BLOOM 7B} & \multicolumn{1}{c|}{Code Instructions} & \textbf{0.174} & \textbf{0.369} & \textbf{0.285} & \textbf{0.150} & \textbf{0.298} & \textbf{0.297} & \textbf{0.232} & \textbf{0.176}\\
\cmidrule(lr){2-10}
 & \begin{tabular}[c]{@{}c@{}}Code Instructions without\\ docstrings and comments\end{tabular} & 0.145 & 0.316 & 0.247 & 0.144 & 0.291 & 0.269 & 0.204 & 0.151\\
 \midrule
\multirow{2}{*}{CodeGen 6B} & \multicolumn{1}{c|}{NL Instructions} & 0.052 & 0.278 & 0.215 & 0.132 & 0.271 & 0.257 & 0.187 & 0.134\\
\cmidrule(lr){2-10}
 & \begin{tabular}[c]{@{}c@{}}NL Instructions with\\ docstrings and comments\end{tabular} & \textbf{0.062} & \textbf{0.312} & \textbf{0.254} & \textbf{0.139} & \textbf{0.293} & \textbf{0.275} & \textbf{0.208} & \textbf{0.148}\\
 \midrule
\multirow{2}{*}{BLOOM 7B} & \multicolumn{1}{c|}{NL Instructions} & \textbf{0.046} & 0.247 & 0.203 & 0.156 & \textbf{0.276} & 0.247 & 0.172 & 0.122\\
\cmidrule(lr){2-10}
 & \begin{tabular}[c]{@{}c@{}}NL Instructions with\\ docstrings and comments\end{tabular} & 0.044 & \textbf{0.303} & \textbf{0.233} & \textbf{0.165} & 0.263 & \textbf{0.266} & \textbf{0.199} & \textbf{0.147} \\
 \bottomrule
\end{tabular}
\caption{Ablation: Zero-Shot Setting. (i) In each model, prompting with pseudo-code instructions results in much higher performance on QA and classification tasks (ii) For each model family, increasing scale helps improve performance (iii) As before, prompting a model designed for code, CodeGen results in better performance than BLOOM. On average, in the CodeGen model, the use of code comments and docstrings helps improve the performance of natural language prompts. However, it appears for BLOOM, only the larger-sized model is able to consistently use the additional details in the prompt to improve performance. }
\label{tab:code-instruct-ablation}
\end{table*}
\subsubsection{Impact of pseudo-code documentation}
In this section, we study the contribution of comments and docstrings present in our pseudo-code instructions towards the improvement in performance. 
We first study the performance of pseudo-code prompts with and without the use of docstrings and code comments.

As can be seen in Table \ref{tab:code-instruct-ablation}, the inclusion of comments as well as the docstring in the pseudo-code instruction prompt helps improve performance. This indicates that not only is the structure of the prompts being exploited by the model, the models are also relying on additional helper text present in the documentation. We, therefore, also investigate if the use of these elements from pseudo-code could also benefit natural language instruction prompts. 

The lower half of table \ref{tab:code-instruct-ablation} studies the performance of natural-language prompts with and without the use of pseudo-code comments and docstring. We find that the performance of natural language instructions also improves by the inclusion of comments and docstring for each model family and configuration. We hypothesize that the gains may be attributable to a form of step-by-step reasoning derived from pseudo-code comments especially in complex tasks. 

\subsection{Summary of findings}
We now summarize our findings for easy reference. 

\noindent{\bf Effect of Prompting Style:}
From Table \ref{tab:zero-shot-main} we observe that 0-shot prompting of pre-trained models with pseudo-code prompts results in better performance than natural language prompts. This is true for both code models and language models. The gains are more pronounced for the code models.

\noindent{\bf Effect of Structure in prompts:} Pseudo-code prompts include many elements such as the function declaration, docstring, comments etc. From Table \ref{tab:struct-prompt-main} we find that while information from the function declaration, and a task-indicative function name help, using the complete pseudo-code prompt is most useful. 

Further, from Table \ref{tab:code-instruct-ablation} we find that the pseudo-code instruction still works better than any prompt created with natural language instructions, even when docstring and comments from pseudo-code are included in the natural language instruction. This suggests the gains from prompting in pseudo-code are not just due to comments and docstrings (which could help reinforce the task instructions), but also due to clearer instructions in pseudo-code.

\noindent{\bf Effect of Model Size:} From Table \ref{tab:zero-shot-main} we find that in 0-shot settings, with the increase in scale, the performance of pseudo-code instructions improves for both model families. However, when using natural language instructions, this is not the case. We hypothesize, that since none of these models are instruction-tuned, larger scales exacerbate the propensity of the models being primed for language completion. 

\noindent{\bf Code vs. Natural Language models:} We find that code models are better suited for exploiting pseudo-code prompts compared to language models. As can be seen from Table \ref{tab:zero-shot-main} (see metrics for `All Tasks'),  the use of natural language instructions on CodeGen results in better performance than their use on BLOOM.


\section{Conclusion and Future Work}
In this paper we presented our work on prompting with pseudo-code instructions. We created a collection of pseudo-code instructions comprising of $132$ NLP tasks from the Super-NaturalInstructions dataset \cite{wang-etal-2022-super}. We evaluated the performance of the following families of models - CodeGen and BLOOM at different model sizes and found that prompting all models with pseudo-code instructions results in significant gains as compared to prompting with NL instructions. 
Our work opens up multiple directions of future work. 
It is interesting to observe that not only do pseudo-code instructions help when used with code models, they also work better on models designed for natural language tasks. In addition, the fact that code models used in our experiments perform better than NL models, even when prompted with natural language instructions, suggests that it could be useful to explore instruction tuning of code models instead of pure NL models for NL applications. Based on the findings of this paper it may also be useful to consider the effects of instruction fine-tuning with pseudo-code instructions as opposed to NL instructions. 

Another aspect worth studying is how traditional chain-of-thought may compare with pseudo-code prompts -- how would reasoning enabled by pseudo-code instructions compare with chain-of-thought reasoning with and without fine-tuning? Further, pseudo-code instructions may not only be used as direct inputs to a model, but they could also be used to create intermediate responses that a model needs to generate prior to returning a response.  

\section*{Limitations}
Our results have been reported on two model families  -- CodeGen and BLOOM at scales of 2-7B parameters. It remains to be seen if our findings would hold at larger model sizes. It is possible that better reasoning enabled by larger model sizes could reduce the benefit of prompting with pseudo-code instructions but we have not investigated this in our work. In addition, our work does not include any multi-lingual NLP tasks -- BLOOM was specifically trained to be able to support multiple languages and it is possible this model design choice could play a role in our findings when we compare code (CodeGen) and NL (BLOOM) models against each other. Moreover, both models have been trained on different datasets and this also affects the intrinsic reasoning capabilities of these models. Lastly, and importantly, the use of pseudo-code for prompting LLMs is limited by the expectation that it requires technical expertise to write them, thus reducing their widespread usage.


\bibliography{anthology,custom}

\begin{thebibliography}{105}
\expandafter\ifx\csname natexlab\endcsname\relax\def\natexlab#1{#1}\fi

\bibitem[{2dot71mily()}]{YTCC}
2dot71mily.
\newblock Youtube captions corrections.
\newblock \url{https://github.com/2dot71mily/youtube_captions_corrections}.

\bibitem[{Almazrouei et~al.(2023)Almazrouei, Alobeidli, Alshamsi, Cappelli, Cojocaru, Debbah, Goffinet, Heslow, Launay, Malartic, Noune, Pannier, and Penedo}]{falcon40b}
Ebtesam Almazrouei, Hamza Alobeidli, Abdulaziz Alshamsi, Alessandro Cappelli, Ruxandra Cojocaru, Merouane Debbah, Etienne Goffinet, Daniel Heslow, Julien Launay, Quentin Malartic, Badreddine Noune, Baptiste Pannier, and Guilherme Penedo. 2023.
\newblock {Falcon-40B}: an open large language model with state-of-the-art performance.

\bibitem[{Alva-Manchego et~al.(2020)Alva-Manchego, Martin, Bordes, Scarton, Sagot, and Specia}]{alva-manchego-etal-2020-asset}
Fernando Alva-Manchego, Louis Martin, Antoine Bordes, Carolina Scarton, Beno{\^\i}t Sagot, and Lucia Specia. 2020.
\newblock \href {https://doi.org/10.18653/v1/2020.acl-main.424} {{ASSET}: {A} dataset for tuning and evaluation of sentence simplification models with multiple rewriting transformations}.
\newblock In \emph{Proceedings of the 58th Annual Meeting of the Association for Computational Linguistics}, pages 4668--4679, Online. Association for Computational Linguistics.

\bibitem[{Aminabadi et~al.(2022)Aminabadi, Rajbhandari, Zhang, Awan, Li, Li, Zheng, Rasley, Smith, Ruwase, and He}]{aminabadi2022deepspeed}
Reza~Yazdani Aminabadi, Samyam Rajbhandari, Minjia Zhang, Ammar~Ahmad Awan, Cheng Li, Du~Li, Elton Zheng, Jeff Rasley, Shaden Smith, Olatunji Ruwase, and Yuxiong He. 2022.
\newblock \href {http://arxiv.org/abs/2207.00032} {Deepspeed inference: Enabling efficient inference of transformer models at unprecedented scale}.

\bibitem[{Arora et~al.(2023)Arora, Narayan, Chen, Orr, Guha, Bhatia, Chami, and Re}]{arora2023ask}
Simran Arora, Avanika Narayan, Mayee~F Chen, Laurel Orr, Neel Guha, Kush Bhatia, Ines Chami, and Christopher Re. 2023.
\newblock \href {https://openreview.net/forum?id=bhUPJnS2g0X} {Ask me anything: A simple strategy for prompting language models}.
\newblock In \emph{The Eleventh International Conference on Learning Representations}.

\bibitem[{Bach et~al.(2022)Bach, Sanh, Yong, Webson, Raffel, Nayak, Sharma, Kim, Bari, Fevry et~al.}]{bach2022promptsource}
Stephen~H Bach, Victor Sanh, Zheng-Xin Yong, Albert Webson, Colin Raffel, Nihal~V Nayak, Abheesht Sharma, Taewoon Kim, M~Saiful Bari, Thibault Fevry, et~al. 2022.
\newblock Promptsource: An integrated development environment and repository for natural language prompts.
\newblock \emph{arXiv preprint arXiv:2202.01279}.

\bibitem[{Bastan et~al.(2020)Bastan, Koupaee, Son, Sicoli, and Balasubramanian}]{bastan2020authors}
Mohaddeseh Bastan, Mahnaz Koupaee, Youngseo Son, Richard Sicoli, and Niranjan Balasubramanian. 2020.
\newblock \href {http://arxiv.org/abs/2011.06128} {Author's sentiment prediction}.

\bibitem[{Ben~Zhou and Roth(2019)}]{ZKNR19}
Qiang~Ning Ben~Zhou, Daniel~Khashabi and Dan Roth. 2019.
\newblock “going on a vacation” takes longer than “going for a walk”: A study of temporal commonsense understanding.
\newblock In \emph{EMNLP}.

\bibitem[{Bisk et~al.(2020)Bisk, Zellers, Bras, Gao, and Choi}]{Bisk2020}
Yonatan Bisk, Rowan Zellers, Ronan~Le Bras, Jianfeng Gao, and Yejin Choi. 2020.
\newblock Piqa: Reasoning about physical commonsense in natural language.
\newblock In \emph{Thirty-Fourth AAAI Conference on Artificial Intelligence}.

\bibitem[{Biten et~al.(2019)Biten, Tito, Mafla, Gomez, Rusinol, Valveny, Jawahar, and Karatzas}]{biten2019scene}
Ali~Furkan Biten, Ruben Tito, Andres Mafla, Lluis Gomez, Mar{\c{c}}al Rusinol, Ernest Valveny, CV~Jawahar, and Dimosthenis Karatzas. 2019.
\newblock Scene text visual question answering.
\newblock In \emph{Proceedings of the IEEE/CVF international conference on computer vision}, pages 4291--4301.

\bibitem[{Bowman et~al.(2015)Bowman, Angeli, Potts, and Manning}]{bowman-etal-2015-large}
Samuel~R. Bowman, Gabor Angeli, Christopher Potts, and Christopher~D. Manning. 2015.
\newblock \href {https://doi.org/10.18653/v1/D15-1075} {A large annotated corpus for learning natural language inference}.
\newblock In \emph{Proceedings of the 2015 Conference on Empirical Methods in Natural Language Processing}, pages 632--642, Lisbon, Portugal. Association for Computational Linguistics.

\bibitem[{Camburu et~al.(2018)Camburu, Rockt"{a}schel, Lukasiewicz, and Blunsom}]{NIPS2018_8163}
Oana-Maria Camburu, Tim Rockt"{a}schel, Thomas Lukasiewicz, and Phil Blunsom. 2018.
\newblock \href {http://papers.nips.cc/paper/8163-e-snli-natural-language-inference-with-natural-language-explanations.pdf} {e-snli: Natural language inference with natural language explanations}.
\newblock In S.~Bengio, H.~Wallach, H.~Larochelle, K.~Grauman, N.~Cesa-Bianchi, and R.~Garnett, editors, \emph{Advances in Neural Information Processing Systems 31}, pages 9539--9549. Curran Associates, Inc.

\bibitem[{Cho and May(2020)}]{cho2020spolin}
Hyundong Cho and Jonathan May. 2020.
\newblock Grounding conversations with improvised dialogues.
\newblock In \emph{Proceedings of the 58th Annual Meeting of the Association for Computational Linguistics}.

\bibitem[{Chowdhery et~al.(2022)Chowdhery, Narang, Devlin, Bosma, Mishra, Roberts, Barham, Chung, Sutton, Gehrmann, Schuh, Shi, Tsvyashchenko, Maynez, Rao, Barnes, Tay, Shazeer, Prabhakaran, Reif, Du, Hutchinson, Pope, Bradbury, Austin, Isard, Gur-Ari, Yin, Duke, Levskaya, Ghemawat, Dev, Michalewski, Garcia, Misra, Robinson, Fedus, Zhou, Ippolito, Luan, Lim, Zoph, Spiridonov, Sepassi, Dohan, Agrawal, Omernick, Dai, Pillai, Pellat, Lewkowycz, Moreira, Child, Polozov, Lee, Zhou, Wang, Saeta, Diaz, Firat, Catasta, Wei, Meier-Hellstern, Eck, Dean, Petrov, and Fiedel}]{chowdhery2022palm}
Aakanksha Chowdhery, Sharan Narang, Jacob Devlin, Maarten Bosma, Gaurav Mishra, Adam Roberts, Paul Barham, Hyung~Won Chung, Charles Sutton, Sebastian Gehrmann, Parker Schuh, Kensen Shi, Sasha Tsvyashchenko, Joshua Maynez, Abhishek Rao, Parker Barnes, Yi~Tay, Noam Shazeer, Vinodkumar Prabhakaran, Emily Reif, Nan Du, Ben Hutchinson, Reiner Pope, James Bradbury, Jacob Austin, Michael Isard, Guy Gur-Ari, Pengcheng Yin, Toju Duke, Anselm Levskaya, Sanjay Ghemawat, Sunipa Dev, Henryk Michalewski, Xavier Garcia, Vedant Misra, Kevin Robinson, Liam Fedus, Denny Zhou, Daphne Ippolito, David Luan, Hyeontaek Lim, Barret Zoph, Alexander Spiridonov, Ryan Sepassi, David Dohan, Shivani Agrawal, Mark Omernick, Andrew~M. Dai, Thanumalayan~Sankaranarayana Pillai, Marie Pellat, Aitor Lewkowycz, Erica Moreira, Rewon Child, Oleksandr Polozov, Katherine Lee, Zongwei Zhou, Xuezhi Wang, Brennan Saeta, Mark Diaz, Orhan Firat, Michele Catasta, Jason Wei, Kathy Meier-Hellstern, Douglas Eck, Jeff Dean, Slav Petrov, and Noah Fiedel. 2022.
\newblock \href {http://arxiv.org/abs/2204.02311} {Palm: Scaling language modeling with pathways}.

\bibitem[{Chung et~al.(2022)Chung, Hou, Longpre, Zoph, Tay, Fedus, Li, Wang, Dehghani, Brahma et~al.}]{chung2022scaling}
Hyung~Won Chung, Le~Hou, Shayne Longpre, Barret Zoph, Yi~Tay, William Fedus, Eric Li, Xuezhi Wang, Mostafa Dehghani, Siddhartha Brahma, et~al. 2022.
\newblock Scaling instruction-finetuned language models.
\newblock \emph{arXiv preprint arXiv:2210.11416}.

\bibitem[{cjadams et~al.(2019)cjadams, Borkan, inversion, Sorensen, Dixon, Vasserman, and nithum}]{jigsaw-unintended-bias-in-toxicity-classification}
cjadams, Daniel Borkan, inversion, Jeffrey Sorensen, Lucas Dixon, Lucy Vasserman, and nithum. 2019.
\newblock \href {https://kaggle.com/competitions/jigsaw-unintended-bias-in-toxicity-classification} {Jigsaw unintended bias in toxicity classification}.

\bibitem[{Clark et~al.(2019)Clark, Lee, Chang, Kwiatkowski, Collins, and Toutanova}]{clark-etal-2019-boolq}
Christopher Clark, Kenton Lee, Ming-Wei Chang, Tom Kwiatkowski, Michael Collins, and Kristina Toutanova. 2019.
\newblock \href {https://doi.org/10.18653/v1/N19-1300} {{B}ool{Q}: Exploring the surprising difficulty of natural yes/no questions}.
\newblock In \emph{Proceedings of the 2019 Conference of the North {A}merican Chapter of the Association for Computational Linguistics: Human Language Technologies, Volume 1 (Long and Short Papers)}, pages 2924--2936, Minneapolis, Minnesota. Association for Computational Linguistics.

\bibitem[{Conneau and Kiela(2018)}]{conneau-kiela-2018-senteval}
Alexis Conneau and Douwe Kiela. 2018.
\newblock \href {https://aclanthology.org/L18-1269} {{S}ent{E}val: An evaluation toolkit for universal sentence representations}.
\newblock In \emph{Proceedings of the Eleventh International Conference on Language Resources and Evaluation ({LREC} 2018)}, Miyazaki, Japan. European Language Resources Association (ELRA).

\bibitem[{Dasigi et~al.(2019)Dasigi, Liu, Marasovi{\'c}, Smith, and Gardner}]{dasigi-etal-2019-quoref}
Pradeep Dasigi, Nelson~F. Liu, Ana Marasovi{\'c}, Noah~A. Smith, and Matt Gardner. 2019.
\newblock \href {https://doi.org/10.18653/v1/D19-1606} {{Q}uoref: A reading comprehension dataset with questions requiring coreferential reasoning}.
\newblock In \emph{Proceedings of the 2019 Conference on Empirical Methods in Natural Language Processing and the 9th International Joint Conference on Natural Language Processing (EMNLP-IJCNLP)}, pages 5925--5932, Hong Kong, China. Association for Computational Linguistics.

\bibitem[{Davidson et~al.(2017)Davidson, Warmsley, Macy, and Weber}]{Davidson_Warmsley_Macy_Weber_2017}
Thomas Davidson, Dana Warmsley, Michael Macy, and Ingmar Weber. 2017.
\newblock \href {https://doi.org/10.1609/icwsm.v11i1.14955} {Automated hate speech detection and the problem of offensive language}.
\newblock \emph{Proceedings of the International AAAI Conference on Web and Social Media}, 11(1):512--515.

\bibitem[{Dua et~al.(2019)Dua, Wang, Dasigi, Stanovsky, Singh, and Gardner}]{Dua2019DROPAR}
Dheeru Dua, Yizhong Wang, Pradeep Dasigi, Gabriel Stanovsky, Sameer Singh, and Matt Gardner. 2019.
\newblock Drop: A reading comprehension benchmark requiring discrete reasoning over paragraphs.
\newblock In \emph{North American Chapter of the Association for Computational Linguistics}.

\bibitem[{Durmus and Cardie(2019)}]{durmus-cardie-2019-corpus}
Esin Durmus and Claire Cardie. 2019.
\newblock \href {https://doi.org/10.18653/v1/P19-1057} {A corpus for modeling user and language effects in argumentation on online debating}.
\newblock In \emph{Proceedings of the 57th Annual Meeting of the Association for Computational Linguistics}, pages 602--607, Florence, Italy. Association for Computational Linguistics.

\bibitem[{Elazar and Goldberg(2019)}]{elazar-goldberg-2019-wheres}
Yanai Elazar and Yoav Goldberg. 2019.
\newblock \href {https://doi.org/10.1162/tacl_a_00280} {Where{'}s my head? {D}efinition, data set, and models for numeric fused-head identification and resolution}.
\newblock \emph{Transactions of the Association for Computational Linguistics}, 7:519--535.

\bibitem[{Face()}]{amazon-polarity-dataset}
Hugging Face.
\newblock Amazon polarity dataset.
\newblock \url{https://huggingface.co/datasets/amazon_polarity}.

\bibitem[{Fu and Khot(2022)}]{fu2022gptroadmap}
Hao Fu, Yao;~Peng and Tushar Khot. 2022.
\newblock \href {https://yaofu.notion.site/How-does-GPT-Obtain-its-Ability-Tracing-Emergent-Abilities-of-Language-Models-to-their-Sources-b9a57ac0fcf74f30a1ab9e3e36fa1dc1} {How does gpt obtain its ability? tracing emergent abilities of language models to their sources}.
\newblock \emph{Yao Fu’s Notion}.

\bibitem[{Gao et~al.(2020)Gao, Biderman, Black, Golding, Hoppe, Foster, Phang, He, Thite, Nabeshima, Presser, and Leahy}]{gao2020pile}
Leo Gao, Stella Biderman, Sid Black, Laurence Golding, Travis Hoppe, Charles Foster, Jason Phang, Horace He, Anish Thite, Noa Nabeshima, Shawn Presser, and Connor Leahy. 2020.
\newblock \href {http://arxiv.org/abs/2101.00027} {The pile: An 800gb dataset of diverse text for language modeling}.

\bibitem[{Geva et~al.(2021)Geva, Khashabi, Segal, Khot, Roth, and Berant}]{geva-etal-2021-aristotle}
Mor Geva, Daniel Khashabi, Elad Segal, Tushar Khot, Dan Roth, and Jonathan Berant. 2021.
\newblock \href {https://doi.org/10.1162/tacl_a_00370} {Did aristotle use a laptop? a question answering benchmark with implicit reasoning strategies}.
\newblock \emph{Transactions of the Association for Computational Linguistics}, 9:346--361.

\bibitem[{Go et~al.(2009)Go, Bhayani, and Huang}]{go2009twitter}
Alec Go, Richa Bhayani, and Lei Huang. 2009.
\newblock Twitter sentiment classification using distant supervision.
\newblock \emph{CS224N project report, Stanford}, 1(12):2009.

\bibitem[{Graff et~al.(2003)Graff, Kong, Chen, and Maeda}]{graff2003english}
David Graff, Junbo Kong, Ke~Chen, and Kazuaki Maeda. 2003.
\newblock English gigaword.
\newblock \emph{Linguistic Data Consortium, Philadelphia}, 4(1):34.

\bibitem[{Henderson et~al.(2014)Henderson, Thomson, and Williams}]{henderson2014third}
Matthew Henderson, Blaise Thomson, and Jason~D Williams. 2014.
\newblock The third dialog state tracking challenge.
\newblock In \emph{2014 IEEE Spoken Language Technology Workshop (SLT)}, pages 324--329. IEEE.

\bibitem[{Hendrycks et~al.(2021)Hendrycks, Burns, Basart, Zou, Mazeika, Song, and Steinhardt}]{hendryckstest2021}
Dan Hendrycks, Collin Burns, Steven Basart, Andy Zou, Mantas Mazeika, Dawn Song, and Jacob Steinhardt. 2021.
\newblock Measuring massive multitask language understanding.
\newblock \emph{Proceedings of the International Conference on Learning Representations (ICLR)}.

\bibitem[{Hossain et~al.(2020)Hossain, Krumm, Gamon, and Kautz}]{hossain-etal-2020-semeval}
Nabil Hossain, John Krumm, Michael Gamon, and Henry Kautz. 2020.
\newblock \href {https://doi.org/10.18653/v1/2020.semeval-1.98} {{S}em{E}val-2020 task 7: Assessing humor in edited news headlines}.
\newblock In \emph{Proceedings of the Fourteenth Workshop on Semantic Evaluation}, pages 746--758, Barcelona (online). International Committee for Computational Linguistics.

\bibitem[{Huang et~al.(2019)Huang, Le~Bras, Bhagavatula, and Choi}]{huang-etal-2019-cosmos}
Lifu Huang, Ronan Le~Bras, Chandra Bhagavatula, and Yejin Choi. 2019.
\newblock \href {https://doi.org/10.18653/v1/D19-1243} {Cosmos {QA}: Machine reading comprehension with contextual commonsense reasoning}.
\newblock In \emph{Proceedings of the 2019 Conference on Empirical Methods in Natural Language Processing and the 9th International Joint Conference on Natural Language Processing (EMNLP-IJCNLP)}, pages 2391--2401, Hong Kong, China. Association for Computational Linguistics.

\bibitem[{Huang et~al.(2020)Huang, Huang, Ding, Hsu, and Giles}]{huang-etal-2020-coda}
Ting-Hao~Kenneth Huang, Chieh-Yang Huang, Chien-Kuang~Cornelia Ding, Yen-Chia Hsu, and C.~Lee Giles. 2020.
\newblock \href {https://aclanthology.org/2020.nlpcovid19-acl.6} {{CODA-19}: Using a non-expert crowd to annotate research aspects on 10,000+ abstracts in the {COVID-19} open research dataset}.
\newblock In \emph{Proceedings of the 1st Workshop on {NLP} for {COVID-19} at {ACL} 2020}, Online. Association for Computational Linguistics.

\bibitem[{Jiang et~al.(2019)Jiang, Wu, and Jiang}]{jiang-etal-2019-freebaseqa}
Kelvin Jiang, Dekun Wu, and Hui Jiang. 2019.
\newblock \href {https://doi.org/10.18653/v1/N19-1028} {{F}reebase{QA}: A new factoid {QA} data set matching trivia-style question-answer pairs with {F}reebase}.
\newblock In \emph{Proceedings of the 2019 Conference of the North {A}merican Chapter of the Association for Computational Linguistics: Human Language Technologies, Volume 1 (Long and Short Papers)}, pages 318--323, Minneapolis, Minnesota. Association for Computational Linguistics.

\bibitem[{Jin and Szolovits(2018)}]{jin-szolovits-2018-pico}
Di~Jin and Peter Szolovits. 2018.
\newblock \href {https://doi.org/10.18653/v1/W18-2308} {{PICO} element detection in medical text via long short-term memory neural networks}.
\newblock In \emph{Proceedings of the {B}io{NLP} 2018 workshop}, pages 67--75, Melbourne, Australia. Association for Computational Linguistics.

\bibitem[{Joshi et~al.(2017)Joshi, Choi, Weld, and Zettlemoyer}]{joshi-etal-2017-triviaqa}
Mandar Joshi, Eunsol Choi, Daniel Weld, and Luke Zettlemoyer. 2017.
\newblock \href {https://doi.org/10.18653/v1/P17-1147} {{T}rivia{QA}: A large scale distantly supervised challenge dataset for reading comprehension}.
\newblock In \emph{Proceedings of the 55th Annual Meeting of the Association for Computational Linguistics (Volume 1: Long Papers)}, pages 1601--1611, Vancouver, Canada. Association for Computational Linguistics.

\bibitem[{Keung et~al.(2020)Keung, Lu, Szarvas, and Smith}]{keung-etal-2020-multilingual}
Phillip Keung, Yichao Lu, Gy{\"o}rgy Szarvas, and Noah~A. Smith. 2020.
\newblock \href {https://doi.org/10.18653/v1/2020.emnlp-main.369} {The multilingual {A}mazon reviews corpus}.
\newblock In \emph{Proceedings of the 2020 Conference on Empirical Methods in Natural Language Processing (EMNLP)}, pages 4563--4568, Online. Association for Computational Linguistics.

\bibitem[{Khashabi et~al.(2018)Khashabi, Chaturvedi, Roth, Upadhyay, and Roth}]{khashabi-etal-2018-looking}
Daniel Khashabi, Snigdha Chaturvedi, Michael Roth, Shyam Upadhyay, and Dan Roth. 2018.
\newblock \href {https://doi.org/10.18653/v1/N18-1023} {Looking beyond the surface: A challenge set for reading comprehension over multiple sentences}.
\newblock In \emph{Proceedings of the 2018 Conference of the North {A}merican Chapter of the Association for Computational Linguistics: Human Language Technologies, Volume 1 (Long Papers)}, pages 252--262, New Orleans, Louisiana. Association for Computational Linguistics.

\bibitem[{Khashabi et~al.(2017)Khashabi, Khot, Sabharwal, and Roth}]{khashabi-etal-2017-learning}
Daniel Khashabi, Tushar Khot, Ashish Sabharwal, and Dan Roth. 2017.
\newblock \href {https://doi.org/10.18653/v1/K17-1010} {Learning what is essential in questions}.
\newblock In \emph{Proceedings of the 21st Conference on Computational Natural Language Learning ({C}o{NLL} 2017)}, pages 80--89, Vancouver, Canada. Association for Computational Linguistics.

\bibitem[{Khot et~al.(2020)Khot, Clark, Guerquin, Jansen, and Sabharwal}]{Khot_Clark_Guerquin_Jansen_Sabharwal_2020}
Tushar Khot, Peter Clark, Michal Guerquin, Peter Jansen, and Ashish Sabharwal. 2020.
\newblock \href {https://doi.org/10.1609/aaai.v34i05.6319} {Qasc: A dataset for question answering via sentence composition}.
\newblock \emph{Proceedings of the AAAI Conference on Artificial Intelligence}, 34(05):8082--8090.

\bibitem[{Kojima et~al.(2022)Kojima, Gu, Reid, Matsuo, and Iwasawa}]{stepbystep}
Takeshi Kojima, Shixiang~Shane Gu, Machel Reid, Yutaka Matsuo, and Yusuke Iwasawa. 2022.
\newblock Large language models are zero-shot reasoners.
\newblock \emph{ArXiv}, abs/2205.11916.

\bibitem[{Kwiatkowski et~al.(2019)Kwiatkowski, Palomaki, Redfield, Collins, Parikh, Alberti, Epstein, Polosukhin, Kelcey, Devlin, Lee, Toutanova, Jones, Chang, Dai, Uszkoreit, Le, and Petrov}]{47761}
Tom Kwiatkowski, Jennimaria Palomaki, Olivia Redfield, Michael Collins, Ankur Parikh, Chris Alberti, Danielle Epstein, Illia Polosukhin, Matthew Kelcey, Jacob Devlin, Kenton Lee, Kristina~N. Toutanova, Llion Jones, Ming-Wei Chang, Andrew Dai, Jakob Uszkoreit, Quoc Le, and Slav Petrov. 2019.
\newblock Natural questions: a benchmark for question answering research.
\newblock \emph{Transactions of the Association of Computational Linguistics}.

\bibitem[{Lauren\c{c}on et~al.(2022)Lauren\c{c}on, Saulnier, Wang, Akiki, Villanova~del Moral, Le~Scao, Von~Werra, Mou, Gonz\'{a}lez~Ponferrada, Nguyen, Frohberg, \v{S}a\v{s}ko, Lhoest, McMillan-Major, Dupont, Biderman, Rogers, Ben~allal, De~Toni, Pistilli, Nguyen, Nikpoor, Masoud, Colombo, de~la Rosa, Villegas, Thrush, Longpre, Nagel, Weber, Mu\~{n}oz, Zhu, Van~Strien, Alyafeai, Almubarak, Vu, Gonzalez-Dios, Soroa, Lo, Dey, Ortiz~Suarez, Gokaslan, Bose, Adelani, Phan, Tran, Yu, Pai, Chim, Lepercq, Ilic, Mitchell, Luccioni, and Jernite}]{ROOTS}
Hugo Lauren\c{c}on, Lucile Saulnier, Thomas Wang, Christopher Akiki, Albert Villanova~del Moral, Teven Le~Scao, Leandro Von~Werra, Chenghao Mou, Eduardo Gonz\'{a}lez~Ponferrada, Huu Nguyen, J\"{o}rg Frohberg, Mario \v{S}a\v{s}ko, Quentin Lhoest, Angelina McMillan-Major, Gerard Dupont, Stella Biderman, Anna Rogers, Loubna Ben~allal, Francesco De~Toni, Giada Pistilli, Olivier Nguyen, Somaieh Nikpoor, Maraim Masoud, Pierre Colombo, Javier de~la Rosa, Paulo Villegas, Tristan Thrush, Shayne Longpre, Sebastian Nagel, Leon Weber, Manuel Mu\~{n}oz, Jian Zhu, Daniel Van~Strien, Zaid Alyafeai, Khalid Almubarak, Minh~Chien Vu, Itziar Gonzalez-Dios, Aitor Soroa, Kyle Lo, Manan Dey, Pedro Ortiz~Suarez, Aaron Gokaslan, Shamik Bose, David Adelani, Long Phan, Hieu Tran, Ian Yu, Suhas Pai, Jenny Chim, Violette Lepercq, Suzana Ilic, Margaret Mitchell, Sasha~Alexandra Luccioni, and Yacine Jernite. 2022.
\newblock \href {https://proceedings.neurips.cc/paper_files/paper/2022/file/ce9e92e3de2372a4b93353eb7f3dc0bd-Paper-Datasets_and_Benchmarks.pdf} {The bigscience roots corpus: A 1.6tb composite multilingual dataset}.
\newblock In \emph{Advances in Neural Information Processing Systems}, volume~35, pages 31809--31826. Curran Associates, Inc.

\bibitem[{Levy et~al.(2017)Levy, Seo, Choi, and Zettlemoyer}]{levy-etal-2017-zero}
Omer Levy, Minjoon Seo, Eunsol Choi, and Luke Zettlemoyer. 2017.
\newblock \href {https://doi.org/10.18653/v1/K17-1034} {Zero-shot relation extraction via reading comprehension}.
\newblock In \emph{Proceedings of the 21st Conference on Computational Natural Language Learning ({C}o{NLL} 2017)}, pages 333--342, Vancouver, Canada. Association for Computational Linguistics.

\bibitem[{Li and Roth(2002)}]{li-roth-2002-learning}
Xin Li and Dan Roth. 2002.
\newblock \href {https://aclanthology.org/C02-1150} {Learning question classifiers}.
\newblock In \emph{{COLING} 2002: The 19th International Conference on Computational Linguistics}.

\bibitem[{Li et~al.(2017)Li, Su, Shen, Li, Cao, and Niu}]{li-etal-2017-dailydialog}
Yanran Li, Hui Su, Xiaoyu Shen, Wenjie Li, Ziqiang Cao, and Shuzi Niu. 2017.
\newblock \href {https://aclanthology.org/I17-1099} {{D}aily{D}ialog: A manually labelled multi-turn dialogue dataset}.
\newblock In \emph{Proceedings of the Eighth International Joint Conference on Natural Language Processing (Volume 1: Long Papers)}, pages 986--995, Taipei, Taiwan. Asian Federation of Natural Language Processing.

\bibitem[{Lin et~al.(2019{\natexlab{a}})Lin, Tafjord, Clark, and Gardner}]{Lin2019ReasoningOP}
Kevin Lin, Oyvind Tafjord, Peter Clark, and Matt Gardner. 2019{\natexlab{a}}.
\newblock Reasoning over paragraph effects in situations.
\newblock In \emph{MRQA@EMNLP}.

\bibitem[{Lin et~al.(2019{\natexlab{b}})Lin, Tafjord, Clark, and Gardner}]{lin-etal-2019-reasoning}
Kevin Lin, Oyvind Tafjord, Peter Clark, and Matt Gardner. 2019{\natexlab{b}}.
\newblock \href {https://doi.org/10.18653/v1/D19-5808} {Reasoning over paragraph effects in situations}.
\newblock In \emph{Proceedings of the 2nd Workshop on Machine Reading for Question Answering}, pages 58--62, Hong Kong, China. Association for Computational Linguistics.

\bibitem[{Liu et~al.(2023)Liu, Yuan, Fu, Jiang, Hayashi, and Neubig}]{10.1145/3560815}
Pengfei Liu, Weizhe Yuan, Jinlan Fu, Zhengbao Jiang, Hiroaki Hayashi, and Graham Neubig. 2023.
\newblock \href {https://doi.org/10.1145/3560815} {Pre-train, prompt, and predict: A systematic survey of prompting methods in natural language processing}.
\newblock \emph{ACM Comput. Surv.}, 55(9).

\bibitem[{Longpre et~al.(2023)Longpre, Hou, Vu, Webson, Chung, Tay, Zhou, Le, Zoph, Wei et~al.}]{longpre2023flan}
Shayne Longpre, Le~Hou, Tu~Vu, Albert Webson, Hyung~Won Chung, Yi~Tay, Denny Zhou, Quoc~V Le, Barret Zoph, Jason Wei, et~al. 2023.
\newblock The flan collection: Designing data and methods for effective instruction tuning.
\newblock \emph{arXiv preprint arXiv:2301.13688}.

\bibitem[{Lourie et~al.(2021)Lourie, Le~Bras, and Choi}]{Lourie_Le_Bras_Choi_2021}
Nicholas Lourie, Ronan Le~Bras, and Yejin Choi. 2021.
\newblock \href {https://doi.org/10.1609/aaai.v35i15.17589} {Scruples: A corpus of community ethical judgments on 32,000 real-life anecdotes}.
\newblock \emph{Proceedings of the AAAI Conference on Artificial Intelligence}, 35(15):13470--13479.

\bibitem[{MacCartney and Manning(2007)}]{maccartney2007natural}
Bill MacCartney and Christopher~D Manning. 2007.
\newblock Natural logic for textual inference.
\newblock In \emph{Proceedings of the ACL-PASCAL Workshop on Textual Entailment and Paraphrasing}, pages 193--200.

\bibitem[{Madaan et~al.(2022)Madaan, Zhou, Alon, Yang, and Neubig}]{madaan2022language}
Aman Madaan, Shuyan Zhou, Uri Alon, Yiming Yang, and Graham Neubig. 2022.
\newblock Language models of code are few-shot commonsense learners.
\newblock \emph{arXiv preprint arXiv:2210.07128}.

\bibitem[{MarvinAI()}]{askmarvin}
MarvinAI. 30 March 2023.
\newblock Marvinai.
\newblock \url{https://www.askmarvin.ai/}.

\bibitem[{McCann et~al.(2019)McCann, Keskar, Xiong, and Socher}]{mccann2019the}
Bryan McCann, Nitish~Shirish Keskar, Caiming Xiong, and Richard Socher. 2019.
\newblock \href {https://openreview.net/forum?id=B1lfHhR9tm} {The natural language decathlon: Multitask learning as question answering}.

\bibitem[{Min et~al.(2020)Min, Michael, Hajishirzi, and Zettlemoyer}]{min-etal-2020-ambigqa}
Sewon Min, Julian Michael, Hannaneh Hajishirzi, and Luke Zettlemoyer. 2020.
\newblock \href {https://doi.org/10.18653/v1/2020.emnlp-main.466} {{A}mbig{QA}: Answering ambiguous open-domain questions}.
\newblock In \emph{Proceedings of the 2020 Conference on Empirical Methods in Natural Language Processing (EMNLP)}, pages 5783--5797, Online. Association for Computational Linguistics.

\bibitem[{Mishra et~al.(2021)Mishra, Khashabi, Baral, and Hajishirzi}]{mishra2021cross}
Swaroop Mishra, Daniel Khashabi, Chitta Baral, and Hannaneh Hajishirzi. 2021.
\newblock Cross-task generalization via natural language crowdsourcing instructions.
\newblock \emph{arXiv preprint arXiv:2104.08773}.

\bibitem[{Misra et~al.(2016)Misra, Ecker, and Walker}]{misra-etal-2016-measuring}
Amita Misra, Brian Ecker, and Marilyn Walker. 2016.
\newblock \href {https://doi.org/10.18653/v1/W16-3636} {Measuring the similarity of sentential arguments in dialogue}.
\newblock In \emph{Proceedings of the 17th Annual Meeting of the Special Interest Group on Discourse and Dialogue}, pages 276--287, Los Angeles. Association for Computational Linguistics.

\bibitem[{Mostafazadeh et~al.(2016)Mostafazadeh, Chambers, He, Parikh, Batra, Vanderwende, Kohli, and Allen}]{mostafazadeh-etal-2016-corpus}
Nasrin Mostafazadeh, Nathanael Chambers, Xiaodong He, Devi Parikh, Dhruv Batra, Lucy Vanderwende, Pushmeet Kohli, and James Allen. 2016.
\newblock \href {https://doi.org/10.18653/v1/N16-1098} {A corpus and cloze evaluation for deeper understanding of commonsense stories}.
\newblock In \emph{Proceedings of the 2016 Conference of the North {A}merican Chapter of the Association for Computational Linguistics: Human Language Technologies}, pages 839--849, San Diego, California. Association for Computational Linguistics.

\bibitem[{Mostafazadeh et~al.(2017)Mostafazadeh, Roth, Louis, Chambers, and Allen}]{mostafazadeh2017lsdsem}
Nasrin Mostafazadeh, Michael Roth, Annie Louis, Nathanael Chambers, and James Allen. 2017.
\newblock Lsdsem 2017 shared task: The story cloze test.
\newblock In \emph{Proceedings of the 2nd Workshop on Linking Models of Lexical, Sentential and Discourse-level Semantics}, pages 46--51.

\bibitem[{Nie et~al.(2020)Nie, Williams, Dinan, Bansal, Weston, and Kiela}]{nie2019adversarial}
Yixin Nie, Adina Williams, Emily Dinan, Mohit Bansal, Jason Weston, and Douwe Kiela. 2020.
\newblock Adversarial nli: A new benchmark for natural language understanding.
\newblock In \emph{Proceedings of the 58th Annual Meeting of the Association for Computational Linguistics}. Association for Computational Linguistics.

\bibitem[{Nijkamp et~al.(2023)Nijkamp, Pang, Hayashi, Tu, Wang, Zhou, Savarese, and Xiong}]{codegen}
Erik Nijkamp, Bo~Pang, Hiroaki Hayashi, Lifu Tu, Huan Wang, Yingbo Zhou, Silvio Savarese, and Caiming Xiong. 2023.
\newblock \href {https://openreview.net/forum?id=iaYcJKpY2B_} {Codegen: An open large language model for code with multi-turn program synthesis}.
\newblock In \emph{The Eleventh International Conference on Learning Representations}.

\bibitem[{OpenAI(2023)}]{openai2023gpt4}
OpenAI. 2023.
\newblock \href {http://arxiv.org/abs/2303.08774} {Gpt-4 technical report}.

\bibitem[{Ostermann et~al.(2018)Ostermann, Modi, Roth, Thater, and Pinkal}]{ostermann-etal-2018-mcscript}
Simon Ostermann, Ashutosh Modi, Michael Roth, Stefan Thater, and Manfred Pinkal. 2018.
\newblock \href {https://aclanthology.org/L18-1564} {{MCS}cript: A novel dataset for assessing machine comprehension using script knowledge}.
\newblock In \emph{Proceedings of the Eleventh International Conference on Language Resources and Evaluation ({LREC} 2018)}, Miyazaki, Japan. European Language Resources Association (ELRA).

\bibitem[{Raffel et~al.(2020)Raffel, Shazeer, Roberts, Lee, Narang, Matena, Zhou, Li, and Liu}]{raffel2020exploring}
Colin Raffel, Noam Shazeer, Adam Roberts, Katherine Lee, Sharan Narang, Michael Matena, Yanqi Zhou, Wei Li, and Peter~J Liu. 2020.
\newblock Exploring the limits of transfer learning with a unified text-to-text transformer.
\newblock \emph{The Journal of Machine Learning Research}, 21(1):5485--5551.

\bibitem[{Raganato et~al.(2020)Raganato, Pasini, Camacho-Collados, and Pilehvar}]{raganato-etal-2020-xl}
Alessandro Raganato, Tommaso Pasini, Jose Camacho-Collados, and Mohammad~Taher Pilehvar. 2020.
\newblock \href {https://doi.org/10.18653/v1/2020.emnlp-main.584} {{XL}-{W}i{C}: A multilingual benchmark for evaluating semantic contextualization}.
\newblock In \emph{Proceedings of the 2020 Conference on Empirical Methods in Natural Language Processing (EMNLP)}, pages 7193--7206, Online. Association for Computational Linguistics.

\bibitem[{Rajpurkar et~al.(2018)Rajpurkar, Jia, and Liang}]{rajpurkar-etal-2018-know}
Pranav Rajpurkar, Robin Jia, and Percy Liang. 2018.
\newblock \href {https://doi.org/10.18653/v1/P18-2124} {Know what you don{'}t know: Unanswerable questions for {SQ}u{AD}}.
\newblock In \emph{Proceedings of the 56th Annual Meeting of the Association for Computational Linguistics (Volume 2: Short Papers)}, pages 784--789, Melbourne, Australia. Association for Computational Linguistics.

\bibitem[{Rajpurkar et~al.(2016)Rajpurkar, Zhang, Lopyrev, and Liang}]{rajpurkar-etal-2016-squad}
Pranav Rajpurkar, Jian Zhang, Konstantin Lopyrev, and Percy Liang. 2016.
\newblock \href {https://doi.org/10.18653/v1/D16-1264} {{SQ}u{AD}: 100,000+ questions for machine comprehension of text}.
\newblock In \emph{Proceedings of the 2016 Conference on Empirical Methods in Natural Language Processing}, pages 2383--2392, Austin, Texas. Association for Computational Linguistics.

\bibitem[{Reddy et~al.(2019)Reddy, Chen, and Manning}]{reddy-etal-2019-coqa}
Siva Reddy, Danqi Chen, and Christopher~D. Manning. 2019.
\newblock \href {https://doi.org/10.1162/tacl_a_00266} {{C}o{QA}: A conversational question answering challenge}.
\newblock \emph{Transactions of the Association for Computational Linguistics}, 7:249--266.

\bibitem[{Reynolds and McDonell(2021)}]{10.1145/3411763.3451760}
Laria Reynolds and Kyle McDonell. 2021.
\newblock \href {https://doi.org/10.1145/3411763.3451760} {Prompt programming for large language models: Beyond the few-shot paradigm}.
\newblock In \emph{Extended Abstracts of the 2021 CHI Conference on Human Factors in Computing Systems}, CHI EA '21, New York, NY, USA. Association for Computing Machinery.

\bibitem[{Rudinger et~al.(2020)Rudinger, Shwartz, Hwang, Bhagavatula, Forbes, Le~Bras, Smith, and Choi}]{rudinger-etal-2020-thinking}
Rachel Rudinger, Vered Shwartz, Jena~D. Hwang, Chandra Bhagavatula, Maxwell Forbes, Ronan Le~Bras, Noah~A. Smith, and Yejin Choi. 2020.
\newblock \href {https://doi.org/10.18653/v1/2020.findings-emnlp.418} {Thinking like a skeptic: Defeasible inference in natural language}.
\newblock In \emph{Findings of the Association for Computational Linguistics: EMNLP 2020}, pages 4661--4675, Online. Association for Computational Linguistics.

\bibitem[{Sakaguchi et~al.(2021)Sakaguchi, Bras, Bhagavatula, and Choi}]{sakaguchi2021winogrande}
Keisuke Sakaguchi, Ronan~Le Bras, Chandra Bhagavatula, and Yejin Choi. 2021.
\newblock Winogrande: An adversarial winograd schema challenge at scale.
\newblock \emph{Communications of the ACM}, 64(9):99--106.

\bibitem[{Sap et~al.(2019)Sap, Le~Bras, Allaway, Bhagavatula, Lourie, Rashkin, Roof, Smith, and Choi}]{Sap2019ATOMICAA}
Maarten Sap, Ronan Le~Bras, Emily Allaway, Chandra Bhagavatula, Nicholas Lourie, Hannah Rashkin, Brendan Roof, Noah~A. Smith, and Yejin Choi. 2019.
\newblock \href {https://doi.org/10.1609/aaai.v33i01.33013027} {Atomic: An atlas of machine commonsense for if-then reasoning}.
\newblock \emph{Proceedings of the AAAI Conference on Artificial Intelligence}, 33(01):3027--3035.

\bibitem[{Scao et~al.(2023)Scao, Fan, Akiki, Pavlick, Ilić, Hesslow, Castagné, Luccioni, Yvon, Gallé, Tow, Rush, Biderman, Webson, Ammanamanchi, Wang, Sagot, Muennighoff, del Moral, Ruwase, Bawden, Bekman, McMillan-Major, Beltagy, Nguyen, Saulnier, Tan, Suarez, Sanh, Laurençon, Jernite, Launay, Mitchell, Raffel, Gokaslan, Simhi, Soroa, Aji, Alfassy, Rogers, Nitzav, Xu, Mou, Emezue, Klamm, Leong, van Strien, Adelani, Radev, Ponferrada, Levkovizh, Kim, Natan, Toni, Dupont, Kruszewski, Pistilli, Elsahar, Benyamina, Tran, Yu, Abdulmumin, Johnson, Gonzalez-Dios, de~la Rosa, Chim, Dodge, Zhu, Chang, Frohberg, Tobing, Bhattacharjee, Almubarak, Chen, Lo, Werra, Weber, Phan, allal, Tanguy, Dey, Muñoz, Masoud, Grandury, Šaško, Huang, Coavoux, Singh, Jiang, Vu, Jauhar, Ghaleb, Subramani, Kassner, Khamis, Nguyen, Espejel, de~Gibert, Villegas, Henderson, Colombo, Amuok, Lhoest, Harliman, Bommasani, López, Ribeiro, Osei, Pyysalo, Nagel, Bose, Muhammad, Sharma, Longpre, Nikpoor, Silberberg, Pai, Zink, Torrent,
  Schick, Thrush, Danchev, Nikoulina, Laippala, Lepercq, Prabhu, Alyafeai, Talat, Raja, Heinzerling, Si, Taşar, Salesky, Mielke, Lee, Sharma, Santilli, Chaffin, Stiegler, Datta, Szczechla, Chhablani, Wang, Pandey, Strobelt, Fries, Rozen, Gao, Sutawika, Bari, Al-shaibani, Manica, Nayak, Teehan, Albanie, Shen, Ben-David, Bach, Kim, Bers, Fevry, Neeraj, Thakker, Raunak, Tang, Yong, Sun, Brody, Uri, Tojarieh, Roberts, Chung, Tae, Phang, Press, Li, Narayanan, Bourfoune, Casper, Rasley, Ryabinin, Mishra, Zhang, Shoeybi, Peyrounette, Patry, Tazi, Sanseviero, von Platen, Cornette, Lavallée, Lacroix, Rajbhandari, Gandhi, Smith, Requena, Patil, Dettmers, Baruwa, Singh, Cheveleva, Ligozat, Subramonian, Névéol, Lovering, Garrette, Tunuguntla, Reiter, Taktasheva, Voloshina, Bogdanov, Winata, Schoelkopf, Kalo, Novikova, Forde, Clive, Kasai, Kawamura, Hazan, Carpuat, Clinciu, Kim, Cheng, Serikov, Antverg, van~der Wal, Zhang, Zhang, Gehrmann, Mirkin, Pais, Shavrina, Scialom, Yun, Limisiewicz, Rieser, Protasov, Mikhailov,
  Pruksachatkun, Belinkov, Bamberger, Kasner, Rueda, Pestana, Feizpour, Khan, Faranak, Santos, Hevia, Unldreaj, Aghagol, Abdollahi, Tammour, HajiHosseini, Behroozi, Ajibade, Saxena, Ferrandis, Contractor, Lansky, David, Kiela, Nguyen, Tan, Baylor, Ozoani, Mirza, Ononiwu, Rezanejad, Jones, Bhattacharya, Solaiman, Sedenko, Nejadgholi, Passmore, Seltzer, Sanz, Dutra, Samagaio, Elbadri, Mieskes, Gerchick, Akinlolu, McKenna, Qiu, Ghauri, Burynok, Abrar, Rajani, Elkott, Fahmy, Samuel, An, Kromann, Hao, Alizadeh, Shubber, Wang, Roy, Viguier, Le, Oyebade, Le, Yang, Nguyen, Kashyap, Palasciano, Callahan, Shukla, Miranda-Escalada, Singh, Beilharz, Wang, Brito, Zhou, Jain, Xu, Fourrier, Periñán, Molano, Yu, Manjavacas, Barth, Fuhrimann, Altay, Bayrak, Burns, Vrabec, Bello, Dash, Kang, Giorgi, Golde, Posada, Sivaraman, Bulchandani, Liu, Shinzato, de~Bykhovetz, Takeuchi, Pàmies, Castillo, Nezhurina, Sänger, Samwald, Cullan, Weinberg, Wolf, Mihaljcic, Liu, Freidank, Kang, Seelam, Dahlberg, Broad, Muellner, Fung,
  Haller, Chandrasekhar, Eisenberg, Martin, Canalli, Su, Su, Cahyawijaya, Garda, Deshmukh, Mishra, Kiblawi, Ott, Sang-aroonsiri, Kumar, Schweter, Bharati, Laud, Gigant, Kainuma, Kusa, Labrak, Bajaj, Venkatraman, Xu, Xu, Xu, Tan, Xie, Ye, Bras, Belkada, and Wolf}]{BLOOM}
Teven~Le Scao, Angela Fan, Christopher Akiki, Ellie Pavlick, Suzana Ilić, Daniel Hesslow, Roman Castagné, Alexandra~Sasha Luccioni, François Yvon, Matthias Gallé, Jonathan Tow, Alexander~M. Rush, Stella Biderman, Albert Webson, Pawan~Sasanka Ammanamanchi, Thomas Wang, Benoît Sagot, Niklas Muennighoff, Albert~Villanova del Moral, Olatunji Ruwase, Rachel Bawden, Stas Bekman, Angelina McMillan-Major, Iz~Beltagy, Huu Nguyen, Lucile Saulnier, Samson Tan, Pedro~Ortiz Suarez, Victor Sanh, Hugo Laurençon, Yacine Jernite, Julien Launay, Margaret Mitchell, Colin Raffel, Aaron Gokaslan, Adi Simhi, Aitor Soroa, Alham~Fikri Aji, Amit Alfassy, Anna Rogers, Ariel~Kreisberg Nitzav, Canwen Xu, Chenghao Mou, Chris Emezue, Christopher Klamm, Colin Leong, Daniel van Strien, David~Ifeoluwa Adelani, Dragomir Radev, Eduardo~González Ponferrada, Efrat Levkovizh, Ethan Kim, Eyal~Bar Natan, Francesco~De Toni, Gérard Dupont, Germán Kruszewski, Giada Pistilli, Hady Elsahar, Hamza Benyamina, Hieu Tran, Ian Yu, Idris Abdulmumin,
  Isaac Johnson, Itziar Gonzalez-Dios, Javier de~la Rosa, Jenny Chim, Jesse Dodge, Jian Zhu, Jonathan Chang, Jörg Frohberg, Joseph Tobing, Joydeep Bhattacharjee, Khalid Almubarak, Kimbo Chen, Kyle Lo, Leandro~Von Werra, Leon Weber, Long Phan, Loubna~Ben allal, Ludovic Tanguy, Manan Dey, Manuel~Romero Muñoz, Maraim Masoud, María Grandury, Mario Šaško, Max Huang, Maximin Coavoux, Mayank Singh, Mike Tian-Jian Jiang, Minh~Chien Vu, Mohammad~A. Jauhar, Mustafa Ghaleb, Nishant Subramani, Nora Kassner, Nurulaqilla Khamis, Olivier Nguyen, Omar Espejel, Ona de~Gibert, Paulo Villegas, Peter Henderson, Pierre Colombo, Priscilla Amuok, Quentin Lhoest, Rheza Harliman, Rishi Bommasani, Roberto~Luis López, Rui Ribeiro, Salomey Osei, Sampo Pyysalo, Sebastian Nagel, Shamik Bose, Shamsuddeen~Hassan Muhammad, Shanya Sharma, Shayne Longpre, Somaieh Nikpoor, Stanislav Silberberg, Suhas Pai, Sydney Zink, Tiago~Timponi Torrent, Timo Schick, Tristan Thrush, Valentin Danchev, Vassilina Nikoulina, Veronika Laippala, Violette
  Lepercq, Vrinda Prabhu, Zaid Alyafeai, Zeerak Talat, Arun Raja, Benjamin Heinzerling, Chenglei Si, Davut~Emre Taşar, Elizabeth Salesky, Sabrina~J. Mielke, Wilson~Y. Lee, Abheesht Sharma, Andrea Santilli, Antoine Chaffin, Arnaud Stiegler, Debajyoti Datta, Eliza Szczechla, Gunjan Chhablani, Han Wang, Harshit Pandey, Hendrik Strobelt, Jason~Alan Fries, Jos Rozen, Leo Gao, Lintang Sutawika, M~Saiful Bari, Maged~S. Al-shaibani, Matteo Manica, Nihal Nayak, Ryan Teehan, Samuel Albanie, Sheng Shen, Srulik Ben-David, Stephen~H. Bach, Taewoon Kim, Tali Bers, Thibault Fevry, Trishala Neeraj, Urmish Thakker, Vikas Raunak, Xiangru Tang, Zheng-Xin Yong, Zhiqing Sun, Shaked Brody, Yallow Uri, Hadar Tojarieh, Adam Roberts, Hyung~Won Chung, Jaesung Tae, Jason Phang, Ofir Press, Conglong Li, Deepak Narayanan, Hatim Bourfoune, Jared Casper, Jeff Rasley, Max Ryabinin, Mayank Mishra, Minjia Zhang, Mohammad Shoeybi, Myriam Peyrounette, Nicolas Patry, Nouamane Tazi, Omar Sanseviero, Patrick von Platen, Pierre Cornette,
  Pierre~François Lavallée, Rémi Lacroix, Samyam Rajbhandari, Sanchit Gandhi, Shaden Smith, Stéphane Requena, Suraj Patil, Tim Dettmers, Ahmed Baruwa, Amanpreet Singh, Anastasia Cheveleva, Anne-Laure Ligozat, Arjun Subramonian, Aurélie Névéol, Charles Lovering, Dan Garrette, Deepak Tunuguntla, Ehud Reiter, Ekaterina Taktasheva, Ekaterina Voloshina, Eli Bogdanov, Genta~Indra Winata, Hailey Schoelkopf, Jan-Christoph Kalo, Jekaterina Novikova, Jessica~Zosa Forde, Jordan Clive, Jungo Kasai, Ken Kawamura, Liam Hazan, Marine Carpuat, Miruna Clinciu, Najoung Kim, Newton Cheng, Oleg Serikov, Omer Antverg, Oskar van~der Wal, Rui Zhang, Ruochen Zhang, Sebastian Gehrmann, Shachar Mirkin, Shani Pais, Tatiana Shavrina, Thomas Scialom, Tian Yun, Tomasz Limisiewicz, Verena Rieser, Vitaly Protasov, Vladislav Mikhailov, Yada Pruksachatkun, Yonatan Belinkov, Zachary Bamberger, Zdeněk Kasner, Alice Rueda, Amanda Pestana, Amir Feizpour, Ammar Khan, Amy Faranak, Ana Santos, Anthony Hevia, Antigona Unldreaj, Arash Aghagol,
  Arezoo Abdollahi, Aycha Tammour, Azadeh HajiHosseini, Bahareh Behroozi, Benjamin Ajibade, Bharat Saxena, Carlos~Muñoz Ferrandis, Danish Contractor, David Lansky, Davis David, Douwe Kiela, Duong~A. Nguyen, Edward Tan, Emi Baylor, Ezinwanne Ozoani, Fatima Mirza, Frankline Ononiwu, Habib Rezanejad, Hessie Jones, Indrani Bhattacharya, Irene Solaiman, Irina Sedenko, Isar Nejadgholi, Jesse Passmore, Josh Seltzer, Julio~Bonis Sanz, Livia Dutra, Mairon Samagaio, Maraim Elbadri, Margot Mieskes, Marissa Gerchick, Martha Akinlolu, Michael McKenna, Mike Qiu, Muhammed Ghauri, Mykola Burynok, Nafis Abrar, Nazneen Rajani, Nour Elkott, Nour Fahmy, Olanrewaju Samuel, Ran An, Rasmus Kromann, Ryan Hao, Samira Alizadeh, Sarmad Shubber, Silas Wang, Sourav Roy, Sylvain Viguier, Thanh Le, Tobi Oyebade, Trieu Le, Yoyo Yang, Zach Nguyen, Abhinav~Ramesh Kashyap, Alfredo Palasciano, Alison Callahan, Anima Shukla, Antonio Miranda-Escalada, Ayush Singh, Benjamin Beilharz, Bo~Wang, Caio Brito, Chenxi Zhou, Chirag Jain, Chuxin Xu,
  Clémentine Fourrier, Daniel~León Periñán, Daniel Molano, Dian Yu, Enrique Manjavacas, Fabio Barth, Florian Fuhrimann, Gabriel Altay, Giyaseddin Bayrak, Gully Burns, Helena~U. Vrabec, Imane Bello, Ishani Dash, Jihyun Kang, John Giorgi, Jonas Golde, Jose~David Posada, Karthik~Rangasai Sivaraman, Lokesh Bulchandani, Lu~Liu, Luisa Shinzato, Madeleine~Hahn de~Bykhovetz, Maiko Takeuchi, Marc Pàmies, Maria~A Castillo, Marianna Nezhurina, Mario Sänger, Matthias Samwald, Michael Cullan, Michael Weinberg, Michiel~De Wolf, Mina Mihaljcic, Minna Liu, Moritz Freidank, Myungsun Kang, Natasha Seelam, Nathan Dahlberg, Nicholas~Michio Broad, Nikolaus Muellner, Pascale Fung, Patrick Haller, Ramya Chandrasekhar, Renata Eisenberg, Robert Martin, Rodrigo Canalli, Rosaline Su, Ruisi Su, Samuel Cahyawijaya, Samuele Garda, Shlok~S Deshmukh, Shubhanshu Mishra, Sid Kiblawi, Simon Ott, Sinee Sang-aroonsiri, Srishti Kumar, Stefan Schweter, Sushil Bharati, Tanmay Laud, Théo Gigant, Tomoya Kainuma, Wojciech Kusa, Yanis Labrak,
  Yash~Shailesh Bajaj, Yash Venkatraman, Yifan Xu, Yingxin Xu, Yu~Xu, Zhe Tan, Zhongli Xie, Zifan Ye, Mathilde Bras, Younes Belkada, and Thomas Wolf. 2023.
\newblock \href {http://arxiv.org/abs/2211.05100} {Bloom: A 176b-parameter open-access multilingual language model}.

\bibitem[{Sheng and Uthus(2020)}]{sheng-uthus-2020-investigating}
Emily Sheng and David Uthus. 2020.
\newblock \href {https://aclanthology.org/2020.gebnlp-1.9} {Investigating societal biases in a poetry composition system}.
\newblock In \emph{Proceedings of the Second Workshop on Gender Bias in Natural Language Processing}, pages 93--106, Barcelona, Spain (Online). Association for Computational Linguistics.

\bibitem[{Socher et~al.(2013)Socher, Perelygin, Wu, Chuang, Manning, Ng, and Potts}]{socher-etal-2013-recursive}
Richard Socher, Alex Perelygin, Jean Wu, Jason Chuang, Christopher~D. Manning, Andrew Ng, and Christopher Potts. 2013.
\newblock \href {https://aclanthology.org/D13-1170} {Recursive deep models for semantic compositionality over a sentiment treebank}.
\newblock In \emph{Proceedings of the 2013 Conference on Empirical Methods in Natural Language Processing}, pages 1631--1642, Seattle, Washington, USA. Association for Computational Linguistics.

\bibitem[{Stanovsky and Hopkins(2018)}]{Stanovsky2016EMNLP}
Gabriel Stanovsky and Mark Hopkins. 2018.
\newblock Spot the odd man out: Exploring the associative power of lexical resources.
\newblock In \emph{Proceedings of the 2018 Conference on Empirical Methods in Natural Language Processing (EMNLP)}, Brussels, Belgium. Association for Computational Linguistics.

\bibitem[{Sun et~al.(2019)Sun, Yu, Chen, Yu, Choi, and Cardie}]{sun-etal-2019-dream}
Kai Sun, Dian Yu, Jianshu Chen, Dong Yu, Yejin Choi, and Claire Cardie. 2019.
\newblock \href {https://doi.org/10.1162/tacl_a_00264} {{DREAM}: A challenge data set and models for dialogue-based reading comprehension}.
\newblock \emph{Transactions of the Association for Computational Linguistics}, 7:217--231.

\bibitem[{Tafjord et~al.(2018)Tafjord, Clark, Gardner, tau Yih, and Sabharwal}]{Tafjord2018QuaRelAD}
Oyvind Tafjord, Peter Clark, Matt Gardner, Wen tau Yih, and Ashish Sabharwal. 2018.
\newblock Quarel: A dataset and models for answering questions about qualitative relationships.
\newblock \emph{ArXiv}, abs/1811.08048.

\bibitem[{Tiedemann(2012)}]{TIEDEMANN12.463}
Jörg Tiedemann. 2012.
\newblock Parallel data, tools and interfaces in opus.
\newblock In \emph{Proceedings of the Eight International Conference on Language Resources and Evaluation (LREC'12)}, Istanbul, Turkey. European Language Resources Association (ELRA).

\bibitem[{Van~Hee et~al.(2018)Van~Hee, Lefever, and Hoste}]{van-hee-etal-2018-semeval}
Cynthia Van~Hee, Els Lefever, and V{\'e}ronique Hoste. 2018.
\newblock \href {https://doi.org/10.18653/v1/S18-1005} {{S}em{E}val-2018 task 3: Irony detection in {E}nglish tweets}.
\newblock In \emph{Proceedings of the 12th International Workshop on Semantic Evaluation}, pages 39--50, New Orleans, Louisiana. Association for Computational Linguistics.

\bibitem[{Vaswani et~al.(2017)Vaswani, Shazeer, Parmar, Uszkoreit, Jones, Gomez, Kaiser, and Polosukhin}]{vaswani2017attention}
Ashish Vaswani, Noam Shazeer, Niki Parmar, Jakob Uszkoreit, Llion Jones, Aidan~N Gomez, {\L}ukasz Kaiser, and Illia Polosukhin. 2017.
\newblock Attention is all you need.
\newblock \emph{Advances in neural information processing systems}, 30.

\bibitem[{Vidgen et~al.(2021)Vidgen, Nguyen, Margetts, Rossini, and Tromble}]{vidgen-etal-2021-introducing}
Bertie Vidgen, Dong Nguyen, Helen Margetts, Patricia Rossini, and Rebekah Tromble. 2021.
\newblock \href {https://doi.org/10.18653/v1/2021.naacl-main.182} {Introducing {CAD}: the contextual abuse dataset}.
\newblock In \emph{Proceedings of the 2021 Conference of the North American Chapter of the Association for Computational Linguistics: Human Language Technologies}, pages 2289--2303, Online. Association for Computational Linguistics.

\bibitem[{Wang et~al.(2022{\natexlab{a}})Wang, Kordi, Mishra, Liu, Smith, Khashabi, and Hajishirzi}]{wang2022self}
Yizhong Wang, Yeganeh Kordi, Swaroop Mishra, Alisa Liu, Noah~A Smith, Daniel Khashabi, and Hannaneh Hajishirzi. 2022{\natexlab{a}}.
\newblock Self-instruct: Aligning language model with self generated instructions.
\newblock \emph{arXiv preprint arXiv:2212.10560}.

\bibitem[{Wang et~al.(2022{\natexlab{b}})Wang, Mishra, Alipoormolabashi, Kordi, Mirzaei, Naik, Ashok, Dhanasekaran, Arunkumar, Stap, Pathak, Karamanolakis, Lai, Purohit, Mondal, Anderson, Kuznia, Doshi, Pal, Patel, Moradshahi, Parmar, Purohit, Varshney, Kaza, Verma, Puri, Karia, Doshi, Sampat, Mishra, Reddy~A, Patro, Dixit, and Shen}]{wang-etal-2022-super}
Yizhong Wang, Swaroop Mishra, Pegah Alipoormolabashi, Yeganeh Kordi, Amirreza Mirzaei, Atharva Naik, Arjun Ashok, Arut~Selvan Dhanasekaran, Anjana Arunkumar, David Stap, Eshaan Pathak, Giannis Karamanolakis, Haizhi Lai, Ishan Purohit, Ishani Mondal, Jacob Anderson, Kirby Kuznia, Krima Doshi, Kuntal~Kumar Pal, Maitreya Patel, Mehrad Moradshahi, Mihir Parmar, Mirali Purohit, Neeraj Varshney, Phani~Rohitha Kaza, Pulkit Verma, Ravsehaj~Singh Puri, Rushang Karia, Savan Doshi, Shailaja~Keyur Sampat, Siddhartha Mishra, Sujan Reddy~A, Sumanta Patro, Tanay Dixit, and Xudong Shen. 2022{\natexlab{b}}.
\newblock \href {https://aclanthology.org/2022.emnlp-main.340} {Super-{N}atural{I}nstructions: Generalization via declarative instructions on 1600+ {NLP} tasks}.
\newblock In \emph{Proceedings of the 2022 Conference on Empirical Methods in Natural Language Processing}, pages 5085--5109, Abu Dhabi, United Arab Emirates. Association for Computational Linguistics.

\bibitem[{Warstadt et~al.(2019)Warstadt, Singh, and Bowman}]{warstadt-etal-2019-neural}
Alex Warstadt, Amanpreet Singh, and Samuel~R. Bowman. 2019.
\newblock \href {https://doi.org/10.1162/tacl_a_00290} {Neural network acceptability judgments}.
\newblock \emph{Transactions of the Association for Computational Linguistics}, 7:625--641.

\bibitem[{Webster et~al.(2018)Webster, Recasens, Axelrod, and Baldridge}]{webster-etal-2018-mind}
Kellie Webster, Marta Recasens, Vera Axelrod, and Jason Baldridge. 2018.
\newblock \href {https://doi.org/10.1162/tacl_a_00240} {Mind the {GAP}: A balanced corpus of gendered ambiguous pronouns}.
\newblock \emph{Transactions of the Association for Computational Linguistics}, 6:605--617.

\bibitem[{Wei et~al.(2022{\natexlab{a}})Wei, Bosma, Zhao, Guu, Yu, Lester, Du, Dai, and Le}]{wei2022finetuned}
Jason Wei, Maarten Bosma, Vincent Zhao, Kelvin Guu, Adams~Wei Yu, Brian Lester, Nan Du, Andrew~M. Dai, and Quoc~V Le. 2022{\natexlab{a}}.
\newblock \href {https://openreview.net/forum?id=gEZrGCozdqR} {Finetuned language models are zero-shot learners}.
\newblock In \emph{International Conference on Learning Representations}.

\bibitem[{Wei et~al.(2021)Wei, Bosma, Zhao, Guu, Yu, Lester, Du, Dai, and Le}]{wei2021finetuned}
Jason Wei, Maarten Bosma, Vincent~Y Zhao, Kelvin Guu, Adams~Wei Yu, Brian Lester, Nan Du, Andrew~M Dai, and Quoc~V Le. 2021.
\newblock Finetuned language models are zero-shot learners.
\newblock \emph{arXiv preprint arXiv:2109.01652}.

\bibitem[{Wei et~al.(2022{\natexlab{b}})Wei, Wang, Schuurmans, Bosma, brian ichter, Xia, Chi, Le, and Zhou}]{cot-few-shot}
Jason Wei, Xuezhi Wang, Dale Schuurmans, Maarten Bosma, brian ichter, Fei Xia, Ed~H. Chi, Quoc~V Le, and Denny Zhou. 2022{\natexlab{b}}.
\newblock \href {https://openreview.net/forum?id=_VjQlMeSB_J} {Chain of thought prompting elicits reasoning in large language models}.
\newblock In \emph{Advances in Neural Information Processing Systems}.

\bibitem[{Weller et~al.(2020)Weller, Lourie, Gardner, and Peters}]{weller-etal-2020-learning}
Orion Weller, Nicholas Lourie, Matt Gardner, and Matthew~E. Peters. 2020.
\newblock \href {https://doi.org/10.18653/v1/2020.emnlp-main.105} {Learning from task descriptions}.
\newblock In \emph{Proceedings of the 2020 Conference on Empirical Methods in Natural Language Processing (EMNLP)}, pages 1361--1375, Online. Association for Computational Linguistics.

\bibitem[{Wieting and Gimpel(2018)}]{wieting-gimpel-2018-paranmt}
John Wieting and Kevin Gimpel. 2018.
\newblock \href {https://doi.org/10.18653/v1/P18-1042} {{P}ara{NMT}-50{M}: Pushing the limits of paraphrastic sentence embeddings with millions of machine translations}.
\newblock In \emph{Proceedings of the 56th Annual Meeting of the Association for Computational Linguistics (Volume 1: Long Papers)}, pages 451--462, Melbourne, Australia. Association for Computational Linguistics.

\bibitem[{Williams et~al.(2018)Williams, Nangia, and Bowman}]{williams-etal-2018-broad}
Adina Williams, Nikita Nangia, and Samuel Bowman. 2018.
\newblock \href {https://doi.org/10.18653/v1/N18-1101} {A broad-coverage challenge corpus for sentence understanding through inference}.
\newblock In \emph{Proceedings of the 2018 Conference of the North {A}merican Chapter of the Association for Computational Linguistics: Human Language Technologies, Volume 1 (Long Papers)}, pages 1112--1122, New Orleans, Louisiana. Association for Computational Linguistics.

\bibitem[{Wolf et~al.(2020)Wolf, Debut, Sanh, Chaumond, Delangue, Moi, Cistac, Rault, Louf, Funtowicz, Davison, Shleifer, von Platen, Ma, Jernite, Plu, Xu, Le~Scao, Gugger, Drame, Lhoest, and Rush}]{wolf-etal-2020-transformers}
Thomas Wolf, Lysandre Debut, Victor Sanh, Julien Chaumond, Clement Delangue, Anthony Moi, Pierric Cistac, Tim Rault, Remi Louf, Morgan Funtowicz, Joe Davison, Sam Shleifer, Patrick von Platen, Clara Ma, Yacine Jernite, Julien Plu, Canwen Xu, Teven Le~Scao, Sylvain Gugger, Mariama Drame, Quentin Lhoest, and Alexander Rush. 2020.
\newblock \href {https://doi.org/10.18653/v1/2020.emnlp-demos.6} {Transformers: State-of-the-art natural language processing}.
\newblock In \emph{Proceedings of the 2020 Conference on Empirical Methods in Natural Language Processing: System Demonstrations}, pages 38--45, Online. Association for Computational Linguistics.

\bibitem[{Wolfson et~al.(2020)Wolfson, Geva, Gupta, Gardner, Goldberg, Deutch, and Berant}]{Wolfson2020Break}
Tomer Wolfson, Mor Geva, Ankit Gupta, Matt Gardner, Yoav Goldberg, Daniel Deutch, and Jonathan Berant. 2020.
\newblock Break it down: A question understanding benchmark.
\newblock \emph{Transactions of the Association for Computational Linguistics}.

\bibitem[{Xiong et~al.(2019)Xiong, Wu, Wang, Kulkarni, Yu, Chang, Guo, and Wang}]{xiong-etal-2019-tweetqa}
Wenhan Xiong, Jiawei Wu, Hong Wang, Vivek Kulkarni, Mo~Yu, Shiyu Chang, Xiaoxiao Guo, and William~Yang Wang. 2019.
\newblock \href {https://doi.org/10.18653/v1/P19-1496} {{TWEETQA}: A social media focused question answering dataset}.
\newblock In \emph{Proceedings of the 57th Annual Meeting of the Association for Computational Linguistics}, pages 5020--5031, Florence, Italy. Association for Computational Linguistics.

\bibitem[{Yin et~al.(2018)Yin, Deng, Chen, Vasilescu, and Neubig}]{yin2018mining}
Pengcheng Yin, Bowen Deng, Edgar Chen, Bogdan Vasilescu, and Graham Neubig. 2018.
\newblock \href {https://doi.org/https://doi.org/10.1145/3196398.3196408} {Learning to mine aligned code and natural language pairs from stack overflow}.
\newblock In \emph{International Conference on Mining Software Repositories}, MSR, pages 476--486. ACM.

\bibitem[{Zamfirescu-Pereira et~al.(2023)Zamfirescu-Pereira, Wong, Hartmann, and Yang}]{10.1145/3544548.3581388}
J.D. Zamfirescu-Pereira, Richmond~Y. Wong, Bjoern Hartmann, and Qian Yang. 2023.
\newblock \href {https://doi.org/10.1145/3544548.3581388} {Why johnny can’t prompt: How non-ai experts try (and fail) to design llm prompts}.
\newblock In \emph{Proceedings of the 2023 CHI Conference on Human Factors in Computing Systems}, CHI '23, New York, NY, USA. Association for Computing Machinery.

\bibitem[{Zhang et~al.(2023{\natexlab{a}})Zhang, Dugan, Xu, and Callison-Burch}]{zhang2023exploring}
Li~Zhang, Liam Dugan, Hainiu Xu, and Chris Callison-Burch. 2023{\natexlab{a}}.
\newblock Exploring the curious case of code prompts.
\newblock \emph{arXiv preprint arXiv:2304.13250}.

\bibitem[{Zhang et~al.(2023{\natexlab{b}})Zhang, Xu, Yang, Zhou, You, Arora, and Callison-Burch}]{zhang2023causal}
Li~Zhang, Hainiu Xu, Yue Yang, Shuyan Zhou, Weiqiu You, Manni Arora, and Chris Callison-Burch. 2023{\natexlab{b}}.
\newblock Causal reasoning of entities and events in procedural texts.
\newblock \emph{arXiv preprint arXiv:2301.10896}.

\bibitem[{Zhang et~al.(2018)Zhang, Liu, Liu, Gao, Duh, and Durme}]{Zhang2018ReCoRDBT}
Sheng Zhang, Xiaodong Liu, Jingjing Liu, Jianfeng Gao, Kevin Duh, and Benjamin~Van Durme. 2018.
\newblock Record: Bridging the gap between human and machine commonsense reading comprehension.
\newblock \emph{ArXiv}, abs/1810.12885.

\bibitem[{Zhang et~al.(2019)Zhang, Baldridge, and He}]{paws2019naacl}
Yuan Zhang, Jason Baldridge, and Luheng He. 2019.
\newblock {PAWS: Paraphrase Adversaries from Word Scrambling}.
\newblock In \emph{Proc. of NAACL}.

\bibitem[{Zhao et~al.(2021)Zhao, Wallace, Feng, Klein, and Singh}]{pmlr-v139-zhao21c}
Zihao Zhao, Eric Wallace, Shi Feng, Dan Klein, and Sameer Singh. 2021.
\newblock \href {http://proceedings.mlr.press/v139/zhao21c.html} {Calibrate before use: Improving few-shot performance of language models}.
\newblock In \emph{Proceedings of ICML}, pages 12697--12706.

\bibitem[{Zhong et~al.(2020)Zhong, Zhang, Wang, Liu, and Miao}]{zhong-etal-2020-towards}
Peixiang Zhong, Chen Zhang, Hao Wang, Yong Liu, and Chunyan Miao. 2020.
\newblock \href {https://doi.org/10.18653/v1/2020.emnlp-main.531} {Towards persona-based empathetic conversational models}.
\newblock In \emph{Proceedings of the 2020 Conference on Empirical Methods in Natural Language Processing (EMNLP)}, pages 6556--6566, Online. Association for Computational Linguistics.

\end{thebibliography}
\bibliographystyle{acl_natbib}

\appendix

\label{sec:appendix}

\section{Appendix}

\begin{table*}[!htb]
\centering
\fontsize{6.8}{7}\selectfont
\begin{tabular}{cc|ccc|c|c|ccc}
\toprule
\textbf{Model} & \textbf{\begin{tabular}[c]{@{}c@{}}Instruction\\  Format\end{tabular}} & \multicolumn{3}{c|}{\textbf{Classification Tasks}}      & \textbf{QA Tasks} & \textbf{\begin{tabular}[c]{@{}c@{}}Generation tasks\end{tabular}} & \multicolumn{3}{c}{\textbf{All Tasks}} \\ \midrule
      & & \multicolumn{1}{c}{Macro F1}       & \multicolumn{1}{c}{Micro F1}       & Weighted F1    & ROUGE-L    & ROUGE-L   & \multicolumn{1}{c}{ROUGE-L} & \multicolumn{1}{c}{ANLS}  & EM    \\ \midrule
\multicolumn{1}{c}{Majority Class} &  & \textbf{0.296} & \textbf{0.509} & \textbf{0.362} & - & - & - & - & - \\ \midrule

\multirow{3}{*}{Falcon 7B} & \multicolumn{1}{c|}{Code Instructions} & \textbf{0.068} & \textbf{0.339} & \textbf{0.259} & 0.152 & 0.265 & \textbf{0.275} & \textbf{0.207} & \textbf{0.161} \\
\cmidrule(lr){2-10}

& \begin{tabular}[c]{@{}c@{}}NL Instructions \end{tabular} & 0.017 & 0.206 & 0.197 & \textbf{0.172} & \textbf{0.273} & 0.242 & 0.149 & 0.102 \\

\bottomrule
\end{tabular}
\caption{Performance of models when prompted using pseudo-code instructions and natural language instructions in 0-shot settings. (i) In each model, prompting with pseudo-code instructions results in much higher performance in almost all the tasks}
\label{tab:zero-shot-appendix}
\end{table*}

\subsection{Results on Various LLMs}
We also perform experiments using Falcon-7B \cite{falcon40b} model. The results are presented in Table \ref{tab:zero-shot-appendix}.

\subsection{Pseudo-Code Validation}
To ensure that the pseudo-code instructions follow the guidelines provided, we run an automatic test. The test code calls the \texttt{preprocess} function defined for each example from the Super-NaturalInstructions dataset \cite{wang-etal-2022-super} for that task. The returned values from the \texttt{preprocess} function are compared against the arguments in the function prototype. Any mismatch in the data type or the number of arguments results in error. The instruction creator is given feedback to correct the errors.

\subsubsection{Prompt Styles}
In this section, we describe the various prompting styles used to study the effect of pseudo-code vs NL prompting. Here, we show a simple task to generate the sentiment of a given sentence. This is task $833$ in Super-NaturalInstructions dataset.


\subsubsection{Prompting with Pseudo-code instructions}

\begin{listing}[!htb]
\begin{minted}{python}
def generate_sentiment(sentence: str) -> str:
    """For the given sentence, the task is to
    predict the sentiment. For positive sentiment
    return "positive" else return "negative".

    Parameters:
        sentence (str): input sentence
    Returns:
        str: sentiment of the input
    """

    # predict the sentiment
    if sentiment_is_positive(sentence):
        return "positive"
    else:
        return "negative"

>>> generate_sentiment(
    "that has a charmingly bourbon air."
)
\end{minted}
\caption{Code instructions (0-shot prompt) for sentiment classification task}
\label{example-type:code-instructions-0}
\end{listing}

\begin{listing}[!htb]
\begin{minted}{python}
def generate_sentiment(sentence: str) -> str:
    """For the given sentence, the task is to
    predict the sentiment. For positive sentiment
    return "positive" else return "negative".

    Parameters:
        sentence (str): input sentence
    Returns:
        str: sentiment of the input
    """

    # predict the sentiment
    if sentiment_is_positive(sentence):
        return "positive"
    else:
        return "negative"

>>> generate_sentiment(
    "tormented by the quickened blood of the "
    "roots"
)
"negative"

>>> generate_sentiment(
    "radiant as moses from the mount, he stood"
)
"positive"

>>> generate_sentiment(
    "that has a charmingly bourbon air."
)
\end{minted}
\caption{Code instructions (2-shot prompt) for sentiment classification task}
\label{example-type:code-instructions-2}
\end{listing}

For the pseudo-code prompting, we use the instructions that are created by the authors of this paper. The pseudo-code instructions have a much richer structure than natural language instructions and are more elaborate and simple to understand. They contain docstrings, return types and might also contain comments, function invocations etc. For preparing the few shot examples and the input query, we treat the example as a python interpreter running in the linux terminal and use the special markers `>{}>{}>' for the input. We don't use any special markers for the outputs. An example for 0-shot and 2-shot shot prompting is shown in Listings \ref{example-type:code-instructions-0} and \ref{example-type:code-instructions-2} respectively.

\begin{listing}[!htb]
\begin{minted}{python}
def generate_sentiment(sentence: str) -> str:
    if sentiment_is_positive(sentence):
        return "positive"
    else:
        return "negative"

>>> generate_sentiment(
    "that has a charmingly bourbon air."
)
\end{minted}
\caption{Code instructions without docstrings and comments (0-shot prompt) for sentiment classification task}
\label{example-type:code-instructions-without-docstrings-0}
\end{listing}

\begin{listing}[!htb]
\begin{minted}{python}
def generate_sentiment(sentence: str) -> str:
    if sentiment_is_positive(sentence):
        return "positive"
    else:
        return "negative"

>>> generate_sentiment(
    "tormented by the quickened blood of the "
    "roots"
)
"negative"

>>> generate_sentiment(
    "radiant as moses from the mount, he stood"
)
"positive"

>>> generate_sentiment(
    "that has a charmingly bourbon air."
)
\end{minted}
\caption{Code instructions without docstrings and comments (2-shot prompt) for sentiment classification task}
\label{example-type:code-instructions-without-docstrings-2}
\end{listing}

We also measure the impact of removing the docstrings and comments from the code instruction. An example for 0-shot and 2-shot prompting is shown in Listings \ref{example-type:code-instructions-without-docstrings-0} and \ref{example-type:code-instructions-without-docstrings-2} respectively.

\subsubsection{Prompting with function prototype}

We try prompting the models with function prototypes with all docstrings, comments and code logic removed from the base pseudo-code instruction. The function prototype instructions are composed of the function names, arguments and their types and the return types. This method of prompting is devoid of any pseudo-code. An example for 0-shot and 2-shot prompting is shown in Listings \ref{example-type:function-definition-0} and \ref{example-type:function-definition-2} respectively.

\begin{listing}[!htb]
\begin{minted}{python}
def generate_sentiment(sentence: str) -> str:

>>> generate_sentiment(
    "that has a charmingly bourbon air."
)
\end{minted}
\caption{Function prototype (0-shot prompt) for sentiment classification task}
\label{example-type:function-definition-0}
\end{listing}

\begin{listing}[!htb]
\begin{minted}{python}
def generate_sentiment(sentence: str) -> str:

>>> generate_sentiment(
    "tormented by the quickened blood of the "
    "roots"
)
"negative"

>>> generate_sentiment(
    "radiant as moses from the mount, he stood"
)
"positive"

>>> generate_sentiment(
    "that has a charmingly bourbon air."
)
\end{minted}
\caption{Function prototype (2-shot prompt) for sentiment classification task}
\label{example-type:function-definition-2}
\end{listing}

\subsubsection{Prompting with NL instructions}

For natural language prompts, we use the original instructions provided as part of the Super-NaturalInstructions dataset \cite{wang-etal-2022-super}. For natural language instruction prompting, we use the prompts provided as part of the Super-NaturalInstructions dataset without any modification. We add special `input:' and `output:' markers in the few shot examples and the input query to the model as shown in Listings \ref{example-type:natural-instructions-0} and \ref{example-type:natural-instructions-2}.

\begin{listing}[!htb]
\begin{minted}{text}
In this task, you need to identify the sentiment
of the given sentence as one of "positive" or
"negative".

input: that has a charmingly bourbon air.
output:
\end{minted}
\caption{Natural instructions (0-shot prompt) for sentiment classification task}
\label{example-type:natural-instructions-0}
\end{listing}

\begin{listing}[!htb]
\begin{minted}{text}
In this task, you need to identify the sentiment
of the given sentence as one of "positive" or
"negative".

input: tormented by the quickened blood of the
roots
output: negative

input: radiant as moses from the mount, he stood
output: positive

input: that has a charmingly bourbon air.
output:
\end{minted}
\caption{Natural instructions (2-shot prompt) for sentiment classification task}
\label{example-type:natural-instructions-2}
\end{listing}

\subsubsection{Prompting with NL instructions and NL comments from the pseudo-code}
We also try experimenting by adding the docstrings and comments to the NL instructions from the Super-NaturalInstructions dataset \cite{wang-etal-2022-super} as shown in the example in Listings \ref{example-type:natural-instructions-with-docstrings-0} and \ref{example-type:natural-instructions-with-docstrings-2}.

\begin{listing}[!htb]
\begin{minted}{text}
In this task, you need to identify the sentiment
of the given sentence as one of "positive" or
"negative".

"""For the given sentence, the task is to
predict the sentiment. For positive sentiment
return "positive" else return "negative".

Parameters:
    sentence (str): input sentence
Returns:
    str: sentiment of the input
"""

# predict the sentiment

input: that has a charmingly bourbon air.
output:
\end{minted}
\caption{Natural instructions with docstrings (0-shot prompt) for sentiment classification task}
\label{example-type:natural-instructions-with-docstrings-0}
\end{listing}

\begin{listing}[!htb]
\begin{minted}{text}
In this task, you need to identify the sentiment
of the given sentence as one of "positive" or
"negative".

"""For the given sentence, the task is to
predict the sentiment. For positive sentiment
return "positive" else return "negative".

Parameters:
    sentence (str): input sentence
Returns:
    str: sentiment of the input
"""

# predict the sentiment

input: tormented by the quickened blood of the
roots
output: negative

input: radiant as moses from the mount, he stood
output: positive

input: that has a charmingly bourbon air.
output:
\end{minted}
\caption{Natural instructions with docstrings (2-shot prompt) for sentiment classification task}
\label{example-type:natural-instructions-with-docstrings-2}
\end{listing}


\subsubsection{Prompting without instructions}
We also study the effect of prompting without instructions. We try this method of prompting in three settings:
\begin{enumerate}
    \item Function Invocation (refer Listings \ref{example-type:no-instruction-code-examples-0} and \ref{example-type:no-instruction-code-examples-2})
    \item Generic Invocation (refer Listings \ref{example-type:no-instruction-generic-function-call-0} and \ref{example-type:no-instruction-generic-function-call-2})
    \item Natural Language examples (refer Listings \ref{example-type:no-instruction-natural-examples-0} and \ref{example-type:no-instruction-natural-examples-2})
\end{enumerate}

\begin{listing}[!htb]
\begin{minted}{python}
>>> generate_sentiment(
    "that has a charmingly bourbon air."
)
\end{minted}
\caption{Function invocation (0-shot prompt) for sentiment classification task}
\label{example-type:no-instruction-code-examples-0}
\end{listing}

\begin{listing}[!htb]
\begin{minted}{python}
>>> generate_sentiment(
    "tormented by the quickened blood of the "
    "roots"
)
"negative"

>>> generate_sentiment(
    "radiant as moses from the mount, he stood"
)
"positive"

>>> generate_sentiment(
    "that has a charmingly bourbon air."
)
\end{minted}
\caption{Function invocation (2-shot prompt) for sentiment classification task}
\label{example-type:no-instruction-code-examples-2}
\end{listing}


\begin{listing}[!htb]
\begin{minted}{python}
>>> func(
    "that has a charmingly bourbon air."
)
\end{minted}
\caption{Generic function invocation (0-shot prompt) for sentiment classification task}
\label{example-type:no-instruction-generic-function-call-0}
\end{listing}

\begin{listing}[!htb]
\begin{minted}{python}
>>> func(
    "tormented by the quickened blood of the "
    "roots"
)
"negative"

>>> func(
    "radiant as moses from the mount, he stood"
)
"positive"

>>> func(
    "that has a charmingly bourbon air."
)
\end{minted}
\caption{Generic function invocation (2-shot prompt) for sentiment classification task}
\label{example-type:no-instruction-generic-function-call-2}
\end{listing}


\begin{listing}[!htb]
\begin{minted}{text}
input: that has a charmingly bourbon air.
output:
\end{minted}
\caption{Natural examples (0-shot prompt) for sentiment classification task}
\label{example-type:no-instruction-natural-examples-0}
\end{listing}

\begin{listing}[!htb]
\begin{minted}{text}
input: tormented by the quickened blood of the
roots
output: negative

input: radiant as moses from the mount, he stood
output: positive

input: that has a charmingly bourbon air.
output:
\end{minted}
\caption{Natural examples (2-shot prompt) for sentiment classification task}
\label{example-type:no-instruction-natural-examples-2}
\end{listing}


\subsection{2-shot Prompting with Pseudo-code instructions}

\begin{table*}[!htb]
\centering
\fontsize{6.7}{7}\selectfont
\begin{tabular}{c|c|ccc|c|c|ccc}
              \toprule
               \multicolumn{1}{c|}{\textbf{Model}}              & \multicolumn{1}{c|}{\textbf{Instruction Format}} & \multicolumn{3}{c|}{\textbf{Classification Tasks}}                                                        & \multicolumn{1}{c|}{\textbf{QA Tasks}} & \multicolumn{1}{c|}{\textbf{Generation Tasks}} & \multicolumn{3}{c}{\textbf{All Tasks}}                                                     \\ \midrule
\multicolumn{1}{c|}{}                            & \multicolumn{1}{c|}{}                            & \multicolumn{1}{c|}{Macro F1} & \multicolumn{1}{c|}{Micro F1}       & \multicolumn{1}{c|}{Weighted F1}    & \multicolumn{1}{c|}{ROUGE-L}             & \multicolumn{1}{c|}{ROUGE-L}                     & \multicolumn{1}{c|}{ROUGE-L}          & \multicolumn{1}{c|}{ANLS}           & EM   \\ \midrule
\multirow{2}{*}{CodeGen 2B} & Code Instructions           & \multicolumn{1}{c|}{\textbf{0.137}} & \multicolumn{1}{c|}{\textbf{0.295}} & \textbf{0.272} & \textbf{0.187}    & \textbf{0.299}            & \multicolumn{1}{c|}{\textbf{0.269}} & \multicolumn{1}{c|}{\textbf{0.202}} & \multicolumn{1}{c}{\textbf{0.148}} \\ 
                            & NL Instructions        & \multicolumn{1}{c|}{0.000}           & \multicolumn{1}{c|}{0.004}          & 0.006          & 0.082             & 0.130                     & \multicolumn{1}{c|}{0.071}          & \multicolumn{1}{c|}{0.017}          & \multicolumn{1}{c}{0.006}          \\ \midrule
\multirow{2}{*}{CodeGen 6B} & Code Instructions           & \multicolumn{1}{c|}{\textbf{0.145}} & \multicolumn{1}{c|}{\textbf{0.317}} & \textbf{0.292} & \textbf{0.194}    & \textbf{0.304}            & \multicolumn{1}{c|}{\textbf{0.285}} & \multicolumn{1}{c|}{\textbf{0.219}} & \multicolumn{1}{c}{\textbf{0.159}} \\ 
                            & NL Instructions       & \multicolumn{1}{c|}{0.000}          & \multicolumn{1}{c|}{0.001}          & 0.002          & 0.101             & 0.172                     & \multicolumn{1}{c|}{0.089}          & \multicolumn{1}{c|}{0.024}          & \multicolumn{1}{c}{0.006}          \\ \midrule
\multirow{2}{*}{BLOOM 3B}   & Code Instructions          & \multicolumn{1}{c|}{\textbf{0.086}} & \multicolumn{1}{c|}{\textbf{0.254}} & \textbf{0.227} & \textbf{0.151}    & \textbf{0.248}            & \multicolumn{1}{c|}{\textbf{0.226}} & \multicolumn{1}{c|}{\textbf{0.164}} & \multicolumn{1}{c}{\textbf{0.121}} \\ 
                            & NL Instructions       & \multicolumn{1}{c|}{0.005}          & \multicolumn{1}{c|}{0.060}          & 0.060          & 0.151             & 0.207                     & \multicolumn{1}{c|}{0.140}          & \multicolumn{1}{c|}{0.070}          & \multicolumn{1}{c}{0.038}          \\ \midrule
\multirow{2}{*}{BLOOM 7B}   & Code Instructions          & \multicolumn{1}{c|}{\textbf{0.072}} & \multicolumn{1}{c|}{\textbf{0.250}} & \textbf{0.227} & \textbf{0.191}    & \textbf{0.279}            & \multicolumn{1}{c|}{\textbf{0.250}} & \multicolumn{1}{c|}{\textbf{0.176}} & \multicolumn{1}{c}{\textbf{0.124}} \\ 
                            & NL Instructions       & \multicolumn{1}{c|}{0.000}          & \multicolumn{1}{c|}{0.120}          & 0.014          & 0.137             & 0.186                     & \multicolumn{1}{c|}{0.109}          & \multicolumn{1}{c|}{0.041}          & \multicolumn{1}{c}{0.018}          \\ \bottomrule
\end{tabular}
\caption{Performance with 2-shot prompts. (i) In each model, prompting with pseudo-code instructions results in much higher performance (ii) For each model family, increasing scale helps improve performance (iii) As before, prompting a model designed for code, CodeGen results in better performance than BLOOM. (iv) Surprisingly, as compared to 0-shot prompting (Table \ref{tab:zero-shot-main}), there is a marked drop in performance for all model configurations and all tasks, except in QA tasks, where there is an improvement in performance.
} \label{tab:two-shot-main}
\end{table*}

Given that structured prompts, such as those based on function declarations, benefit from 2-shot prompts, we investigate whether the performance of pseudo-code prompts can be further improved with 2-shot prompts. Table \ref{tab:two-shot-main} reports the performance of both families of models - CodeGen and BLOOM when using pseudo-code prompts and natural language instruction prompts in 2-shot settings. 

Interestingly we find that, as compared to the results reported in Table \ref{tab:zero-shot-main} the performance of each corresponding model-prompt configuration is lower than its 0-shot counterpart. While this may appear surprising, similar findings have been reported in prior work \cite{10.1145/3411763.3451760,zhang2023exploring}. Perhaps the performance in few-shot settings could improve with additional examples, but we do not experiment with more than 2-shot settings due to limitations imposed by the size of input context length available to models. 

After  a study of outputs generated by the models in 2-shot settings, we observe that in many cases, in the absence of extensive task-specific prompt-engineering and output processing, models are likely to generate additional continuation examples instead of solving the task. 
The fact that the pseudo-code prompts perform better indicate that models seem to ``{\em interpret}'' the instructions better in this form.

\subsection{Ablation Experiments}

As can be seen in Table \ref{tab:nl-instruct-ablation} and \ref{tab:code-instruct-ablation-v2}, the inclusion of comments as well as the docstring in the pseudo-code instruction prompt and natural language instructions helps improve performance for smaller models too.

\begin{table*}[!htb]
\fontsize{6.5}{7}\selectfont
\center
\begin{tabular}{c|c|ccc|c|c|ccc}
\toprule
\textbf{Model}              & \textbf{Instruction Format}                                                              & \multicolumn{3}{c|}{\textbf{Classification Tasks}}                                         & \textbf{QA Tasks} & \textbf{Generation Tasks} & \multicolumn{3}{c}{\textbf{All Tasks}}                                                                          \\ \midrule
                            &                                                                                          & \multicolumn{1}{c|}{Macro F1}       & \multicolumn{1}{c|}{Micro F1}       & Weighted F1    & ROUGE-L             & ROUGE-L                     & \multicolumn{1}{c|}{ROUGE-L}          & \multicolumn{1}{c|}{ANLS}           & EM                         \\ \midrule
\multirow{3}{*}{CodeGen 2B} & NL Instructions                                                                          & \multicolumn{1}{c|}{0.068}          & \multicolumn{1}{c|}{0.306}          & 0.239          & \textbf{0.154}    & 0.254                     & \multicolumn{1}{c|}{0.265}          & \multicolumn{1}{c|}{0.195}          & \multicolumn{1}{c}{0.147}          \\ \cmidrule(lr){2-2} 
                            & \begin{tabular}[c]{@{}c@{}}NL Instructions with\\ docstrings and comments\end{tabular} & \multicolumn{1}{c|}{\textbf{0.098}} & \multicolumn{1}{c|}{\textbf{0.349}} & \textbf{0.270} & 0.136             & \textbf{0.258}            & \multicolumn{1}{c|}{\textbf{0.275}} & \multicolumn{1}{c|}{\textbf{0.208}} & \multicolumn{1}{c}{\textbf{0.161}} \\ \midrule 
\multirow{2}{*}{CodeGen 6B} & NL Instructions                                                                          & \multicolumn{1}{c|}{0.052}          & \multicolumn{1}{c|}{0.278}          & 0.215          & 0.132             & 0.271                     & \multicolumn{1}{c|}{0.257}          & \multicolumn{1}{c|}{0.187}          & \multicolumn{1}{c}{0.134}          \\ \cmidrule(lr){2-2}
                            & \begin{tabular}[c]{@{}c@{}}NL Instructions with\\ docstrings and comments\end{tabular} & \multicolumn{1}{c|}{\textbf{0.062}} & \multicolumn{1}{c|}{\textbf{0.312}} & \textbf{0.254} & \textbf{0.139}    & \textbf{0.293}            & \multicolumn{1}{c|}{\textbf{0.275}} & \multicolumn{1}{c|}{\textbf{0.208}} & \multicolumn{1}{c}{\textbf{0.148}} \\ \midrule
\multirow{2}{*}{BLOOM 3B}   & NL Instructions                                                                          & \multicolumn{1}{c|}{\textbf{0.082}} & \multicolumn{1}{c|}{\textbf{0.275}} & \textbf{0.214} & \textbf{0.159}    & \textbf{0.234}            & \multicolumn{1}{c|}{\textbf{0.250}} & \multicolumn{1}{c|}{\textbf{0.180}} & \multicolumn{1}{c}{\textbf{0.132}} \\ \cmidrule(lr){2-2} 
                            & \begin{tabular}[c]{@{}c@{}}NL Instructions with\\ docstrings and comments\end{tabular} & \multicolumn{1}{c|}{0.046}          & \multicolumn{1}{c|}{0.233}          & 0.209          & 0.121             & 0.202                     & \multicolumn{1}{c|}{0.213}          & \multicolumn{1}{c|}{0.146}          & \multicolumn{1}{c}{0.111}          \\ \midrule 
\multirow{2}{*}{BLOOM 7B}   & NL Instructions                                                                          & \multicolumn{1}{c|}{\textbf{0.046}} & \multicolumn{1}{c|}{0.247}          & 0.203          & 0.156             & \textbf{0.276}            & \multicolumn{1}{c|}{0.247}          & \multicolumn{1}{c|}{0.172}          & \multicolumn{1}{c}{0.122}          \\ \cmidrule(lr){2-2} 
                            & \begin{tabular}[c]{@{}c@{}}NL Instructions with\\ docstrings and comments\end{tabular} & \multicolumn{1}{c|}{0.044}          & \multicolumn{1}{c|}{\textbf{0.303}} & \textbf{0.233} & \textbf{0.165}    & 0.263                     & \multicolumn{1}{c|}{\textbf{0.266}} & \multicolumn{1}{c|}{\textbf{0.199}} & \multicolumn{1}{c}{\textbf{0.147}} \\ \bottomrule 
\end{tabular}
\caption{Ablation: On average, in the CodeGen model the use of code comments and docstrings in 0-shot setting helps improve performance of natural language prompts. However, it appears on BLOOM, only the larger sized model is able to consistently use the additional details in the prompt to improve performance.}\label{tab:nl-instruct-ablation}
\end{table*}

\begin{table*}[!htb]
\fontsize{6.5}{7}\selectfont
\center
\begin{tabular}{c|c|ccc|c|c|ccc}
\toprule
\textbf{Model}              & \textbf{Instruction Format}                                                                   & \multicolumn{3}{c|}{\textbf{Classification Tasks}}                                         & \textbf{QA Tasks} & \textbf{Generation Tasks} & \multicolumn{3}{c}{\textbf{All Tasks}}                                                                          \\ \midrule
                            &                                                                                               & \multicolumn{1}{c|}{Macro F1}       & \multicolumn{1}{c|}{Micro F1}       & Weighted F1    & ROUGE-L             & ROUGE-L                     & \multicolumn{1}{c|}{ROUGE-L}          & \multicolumn{1}{c|}{ANLS}           & EM                         \\ \midrule
\multirow{2}{*}{CodeGen 2B} & Code Instructions                                                                             & \multicolumn{1}{c|}{\textbf{0.272}} & \multicolumn{1}{c|}{\textbf{0.417}} & \textbf{0.354} & \textbf{0.175}    & \textbf{0.317}            & \multicolumn{1}{c|}{\textbf{0.330}} & \multicolumn{1}{c|}{\textbf{0.262}} & \multicolumn{1}{c}{\textbf{0.202}} \\ \cmidrule(lr){2-2} 
                            & \begin{tabular}[c]{@{}c@{}}Code Instructions without\\ docstrings and comments\end{tabular} & \multicolumn{1}{c|}{0.241}          & \multicolumn{1}{c|}{0.389}          & 0.337          & 0.159             & 0.305                     & \multicolumn{1}{c|}{0.309}          & \multicolumn{1}{c|}{0.241}          & \multicolumn{1}{c}{0.185}          \\ \midrule
\multirow{2}{*}{CodeGen 6B} & Code Instructions                                                                             & \multicolumn{1}{c|}{\textbf{0.311}} & \multicolumn{1}{c|}{\textbf{0.444}} & \textbf{0.375} & \textbf{0.201}    & \textbf{0.327}            & \multicolumn{1}{c|}{\textbf{0.354}} & \multicolumn{1}{c|}{\textbf{0.283}} & \multicolumn{1}{c}{\textbf{0.218}} \\ \cmidrule(lr){2-2} 
                            & \begin{tabular}[c]{@{}c@{}}Code Instructions without\\ docstrings and comments\end{tabular} & \multicolumn{1}{c|}{0.263}          & \multicolumn{1}{c|}{0.409}          & 0.348          & 0.195             & 0.327                     & \multicolumn{1}{c|}{0.335}          & \multicolumn{1}{c|}{0.266}          & \multicolumn{1}{c}{0.201}          \\ \midrule
\multirow{2}{*}{BLOOM 3B}   & Code Instructions                                                                             & \multicolumn{1}{c|}{\textbf{0.116}} & \multicolumn{1}{c|}{\textbf{0.351}} & \textbf{0.288} & \textbf{0.147}    & \textbf{0.271}            & \multicolumn{1}{c|}{\textbf{0.279}} & \multicolumn{1}{c|}{\textbf{0.215}} & \multicolumn{1}{c}{\textbf{0.165}} \\ \cmidrule(lr){2-2} 
                            & \begin{tabular}[c]{@{}c@{}}Code Instructions without\\ docstrings and comments\end{tabular} & \multicolumn{1}{c|}{0.094}          & \multicolumn{1}{c|}{0.302}          & 0.249          & 0.132             & 0.259                     & \multicolumn{1}{c|}{0.248}          & \multicolumn{1}{c|}{0.117}          & \multicolumn{1}{c}{0.183}          \\ \midrule
\multirow{2}{*}{BLOOM 7B}   & Code Instructions                                                                             & \multicolumn{1}{c|}{\textbf{0.174}} & \multicolumn{1}{c|}{\textbf{0.369}} & \textbf{0.285} & \textbf{0.150}    & \textbf{0.298}            & \multicolumn{1}{c|}{\textbf{0.297}} & \multicolumn{1}{c|}{\textbf{0.232}} & \multicolumn{1}{c}{\textbf{0.176}} \\ \cmidrule(lr){2-2} 
                            & \begin{tabular}[c]{@{}c@{}}Code Instructions without\\ docstrings and comments\end{tabular} & \multicolumn{1}{c|}{0.145}          & \multicolumn{1}{c|}{0.316}          & 0.247          & 0.144             & 0.291                     & \multicolumn{1}{c|}{0.269}          & \multicolumn{1}{c|}{0.204}          & \multicolumn{1}{c}{0.151}          \\ \bottomrule
\end{tabular}
\caption{Ablation: Using 0-shot code instructions without docstrings and comments (i) In each model, prompting with pseudo-code instructions results in much higher performance on QA and classification tasks (ii) For each model family, increasing scale helps improve performance (iii) As before, prompting a model designed for code, CodeGen results in better performance than BLOOM.}\label{tab:code-instruct-ablation-v2}
 \end{table*}

\end{document}